\documentclass{article}

\usepackage{hyperref}

\usepackage[accepted]{icml2023}

\usepackage{graphicx}
\usepackage{mathtools}
\usepackage{url}
\usepackage{booktabs}
\usepackage{amsmath, amssymb, amsthm}
\usepackage{multirow}
\usepackage{paralist}
\usepackage{enumitem}
\usepackage{algorithm}
\usepackage{algorithmic}
\usepackage{natbib}
\usepackage{stfloats}
\usepackage{subcaption}
\usepackage{wrapfig}
\usepackage{bm}
\usepackage{dsfont}
\usepackage{nicefrac}
\usepackage{makecell}

\usepackage[capitalize,noabbrev]{cleveref}

\theoremstyle{plain}

\theoremstyle{definition}

\theoremstyle{remark}

\usepackage[textsize=tiny]{todonotes}

\usepackage[normalem]{ulem}
\Crefname{algocf}{Algorithm}{Algorithms}
\Crefname{equation}{Eq.}{Eq.}
\Crefname{section}{Sec.}{Sec.}

\DeclareMathOperator*{\argmin}{arg\,min}

\newcommand{\mscript}[1]{\text{\scriptsize{#1}}}
\newcommand{\mbf}[1]{\boldsymbol{\mathbf{#1}}}

\newcommand{\modelname}{ECRTM}
\newcommand{\modelfullname}{Embedding Clustering Regularization Topic Model}
\newcommand{\regname}{ECR}
\newcommand{\regfullname}{Embedding Clustering Regularization}

\icmltitlerunning{Effective Neural Topic Modeling with Embedding Clustering Regularization}

\begin{document}

\twocolumn[
\icmltitle{Effective Neural Topic Modeling with Embedding Clustering Regularization}

\icmlsetsymbol{equal}{*}

\begin{icmlauthorlist}
\icmlauthor{Xiaobao Wu}{ntu}
\icmlauthor{Xinshuai Dong}{cmu}
\icmlauthor{Thong Nguyen}{nus}
\icmlauthor{Anh Tuan Luu}{ntu}
\end{icmlauthorlist}

\icmlaffiliation{ntu}{Nanyang Technological University}
\icmlaffiliation{cmu}{Carnegie Mellon University}
\icmlaffiliation{nus}{National University of Singapore}

\icmlcorrespondingauthor{Xiaobao Wu}{xiaobao002@e.ntu.edu.sg}
\icmlcorrespondingauthor{Anh Tuan Luu}{anhtuan.luu@ntu.edu.sg}
\vskip 0.3in
]

\printAffiliationsAndNotice{}  %

\begin{abstract}
    Topic models have been prevalent for decades with various applications.
    However, existing topic models commonly suffer from the notorious topic collapsing: discovered topics semantically collapse towards each other,
    leading to highly repetitive topics, insufficient topic discovery, and damaged model interpretability.
    In this paper,
    we propose a new neural topic model, \modelfullname{} (\modelname{}).
    Besides the existing reconstruction error,
    we propose a novel \regfullname{} (\regname{}),
    which forces each topic embedding to be the center of a separately aggregated word embedding cluster in the semantic space.
    This enables each produced topic to contain distinct word semantics, which alleviates topic collapsing.
    Regularized by \regname{},
    our \modelname{} generates diverse and coherent topics together with high-quality topic distributions of documents.
    Extensive experiments on benchmark datasets demonstrate that \modelname{} 
    effectively addresses the topic collapsing issue and consistently surpasses state-of-the-art baselines
    in terms of topic quality, topic distributions of documents, and downstream classification tasks.

\end{abstract}

\section{Introduction}
    Topic models have achieved great success in document analysis
    via discovering latent semantics.
    They have facilitated various downstream applications \cite{boyd2017applications}, like content recommendation \cite{mcauley2013hidden}, summarization \cite{Ma2012}, and information retrieval \cite{wang2007topical}.
    Current topic models can be roughly classified as two lines: (1) conventional topic models with probabilistic graphical models \cite{blei2003latent} or matrix factorization \cite{kim2015simultaneous,shi2018short} and (2) neural topic models \cite{Miao2016,Miao2017,Srivastava2017,gupta2019document}.

    \begin{table}
    \centering
    \setlength{\tabcolsep}{1.2mm}
    \renewcommand{\arraystretch}{1.3}
    \resizebox{\linewidth}{!}{
    \begin{tabular}{ll}
        \toprule
        \textbf{Topic\#1}: & \uline{just} show \uline{even} \uline{come} time one \uline{good} \uline{really} going know \\
        \textbf{Topic\#2}: & \uline{just} \uline{even} \uline{really} \uline{something} \uline{come} going \uline{like} actually things \uline{get} \\
        \textbf{Topic\#3}: & \uline{just} one \uline{even} \uline{something} \uline{come} way \uline{really} \uline{like} always \uline{good} \\
        \textbf{Topic\#4}: & \uline{just} \uline{get} going \uline{come} \uline{one} know \uline{even} \uline{really} \uline{something} way \\
        \textbf{Topic\#5}: & \uline{just} \uline{like} inside \uline{get} \uline{even} look \uline{come} \uline{one} everything away \\
        \bottomrule
        \end{tabular}%
    }
    \caption{
        Top related words of the discovered topics by NSTM \cite{zhao2020neural} on IMDB.
        These topics
        semantically collapse towards each other with many uninformative and repetitive words. Repetitions are \uline{underlined}.
    }
    \label{tab_motivation}
\end{table}

    \begin{figure}[!t]
    \centering
    \begin{subfigure}[b]{0.25\linewidth}
        \centering
        \includegraphics[width=\linewidth]{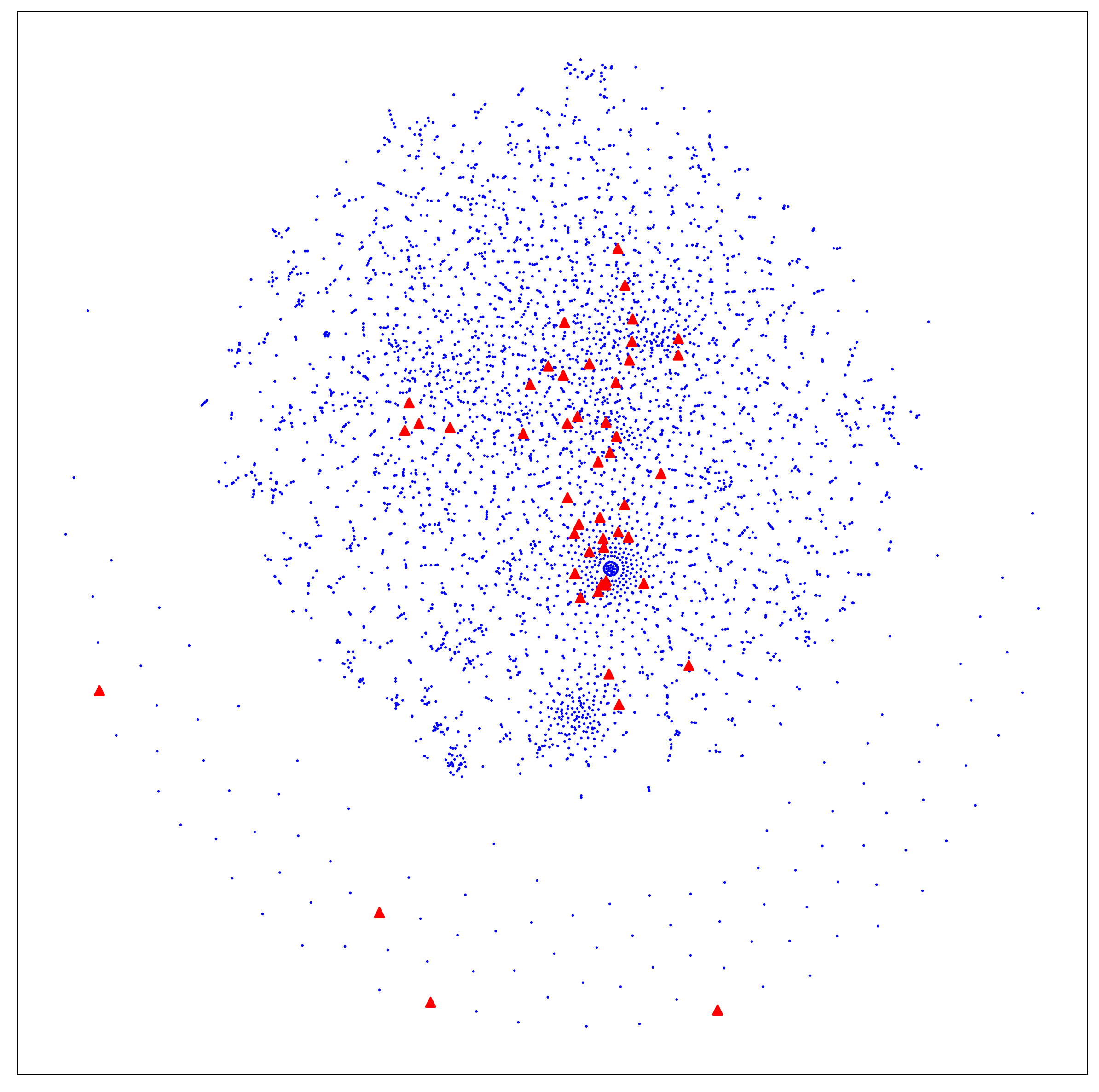}
        \caption{ETM}
        \label{fig_topic_collapse_ETM}
    \end{subfigure}%
    \begin{subfigure}[b]{0.25\linewidth}
        \centering
        \includegraphics[width=\linewidth]{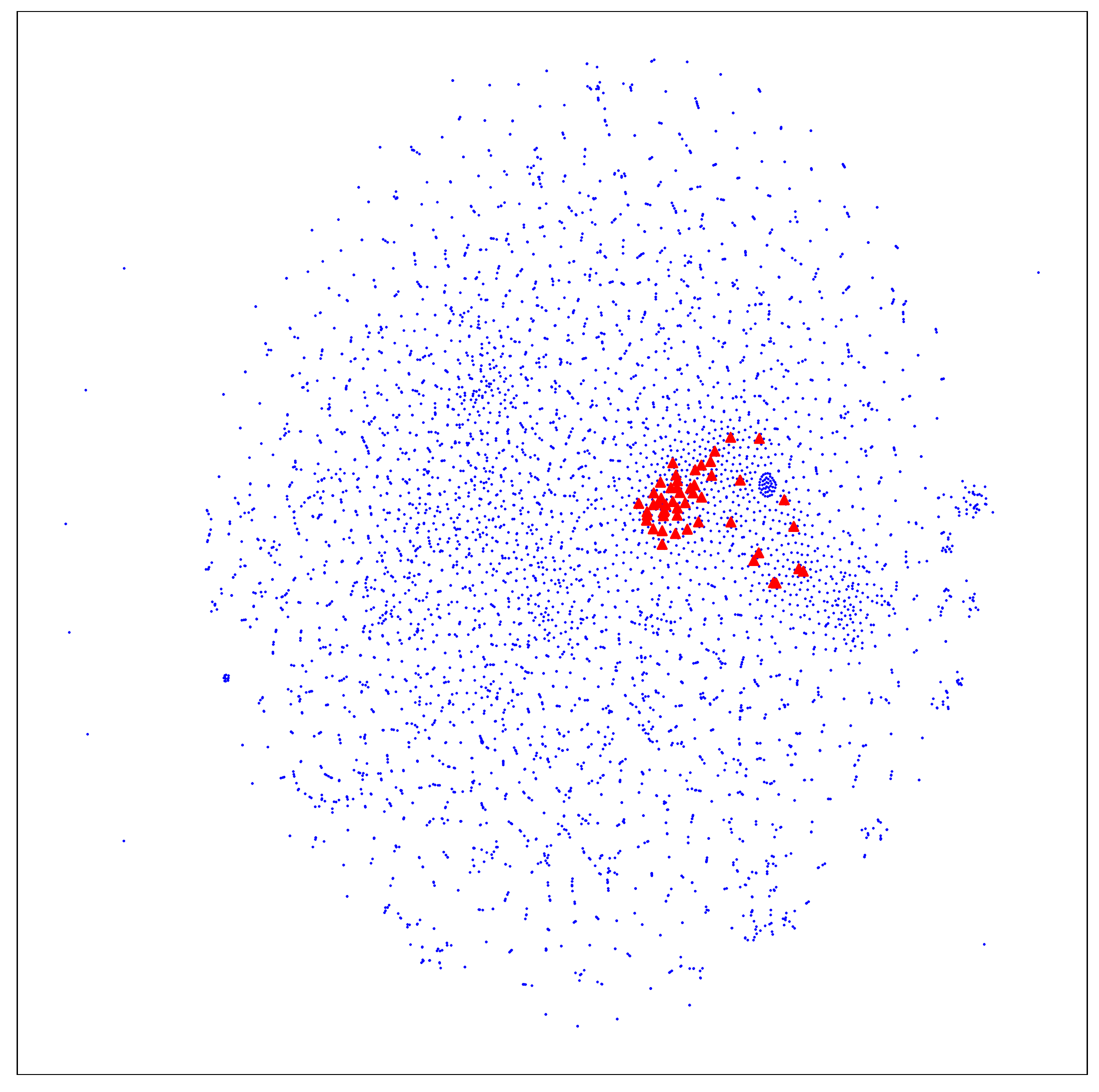}
        \caption{NSTM}
        \label{fig_topic_collapse_NSTM}
    \end{subfigure}%
    \begin{subfigure}[b]{0.25\linewidth}
        \centering
        \includegraphics[width=\linewidth]{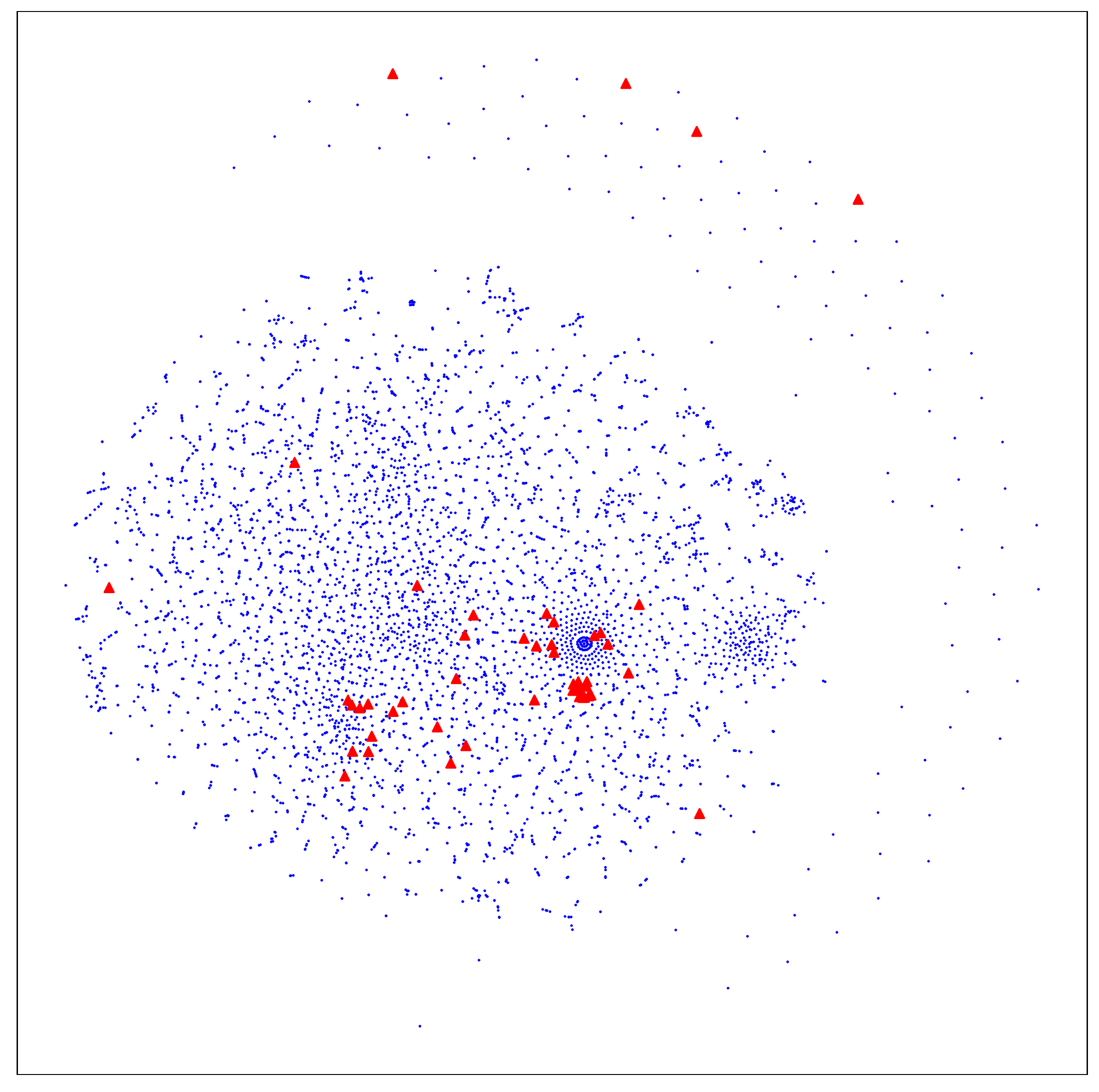}
        \caption{WeTe}
        \label{fig_topic_collapse_WeTe}
    \end{subfigure}%
    \hfill
    \begin{subfigure}[b]{0.25\linewidth}
        \centering
        \includegraphics[width=1.0\linewidth]{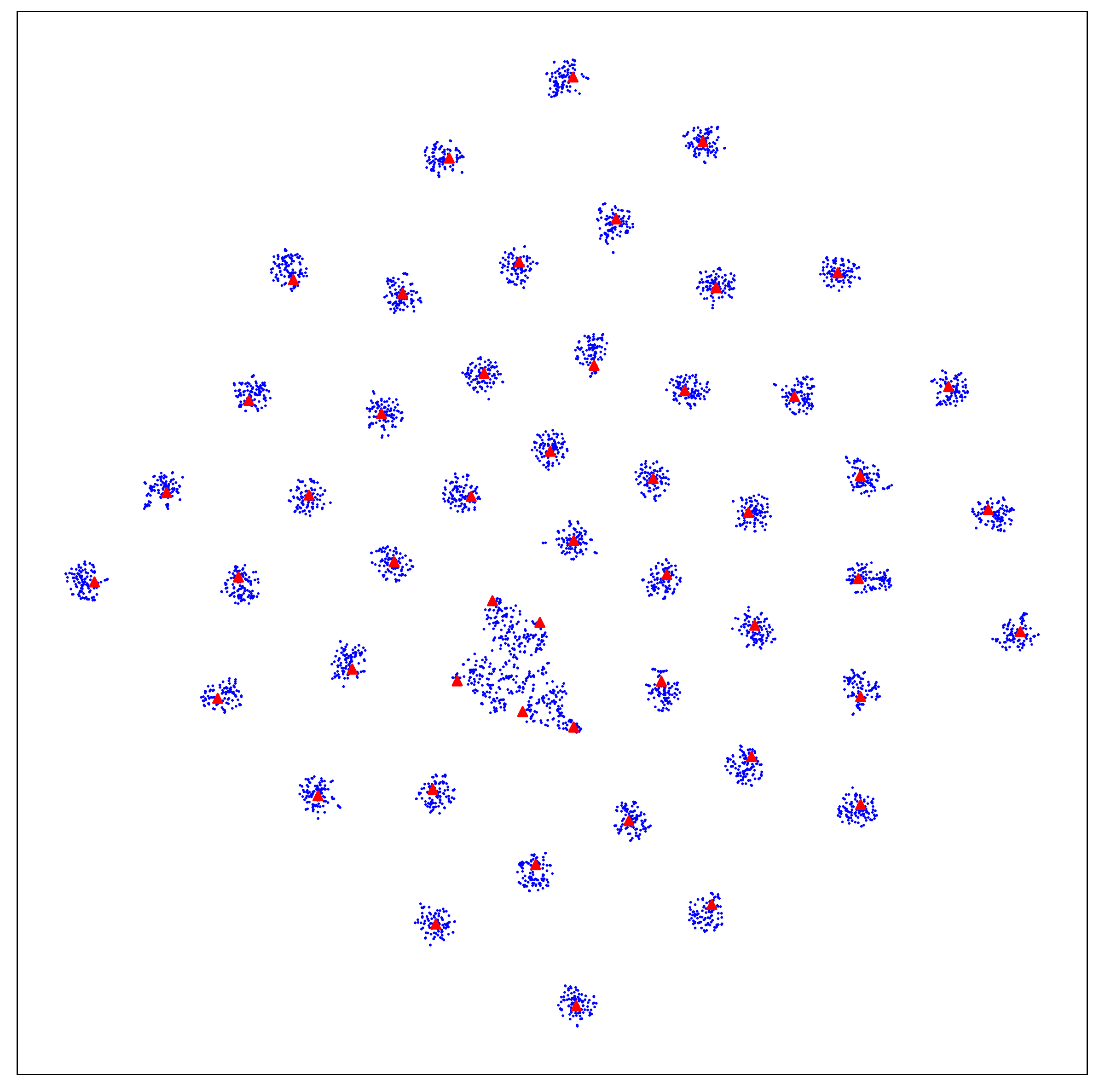}
        \caption{\textbf{\modelname{}}}
        \label{fig_topic_collapse_ECRTM}
    \end{subfigure}%
    \caption{
        t-SNE visualization
        of word embeddings ({\color{blue} $\bullet$}) and topic embeddings ({\color{red} $\blacktriangle$})
        under 50 topics.
        These show while the topic embeddings mostly collapse together in previous state-of-the-art models,
        our \modelname{} successfully avoids the collapsing.
    }
    \label{fig_motivation}
\end{figure}

    However,
    despite the current achievements,
    existing topic models commonly suffer from
    \emph{topic collapsing}: the discovered topics tend to semantically collapse towards each other \cite{Srivastava2017}, as exemplified in \Cref{tab_motivation}.
    We see these collapsed topics include many uninformative and repetitive words.
    This brings about several issues:
    \begin{inparaenum}[(\bgroup\bfseries 1\egroup)]
        \item
            Topic collapsing results in highly repetitive topics, which are less useful for downstream applications \cite{wallach2009rethinking,Nan2019}.
        \item
            Topic collapsing incurs insufficient topic discovery.
            Many latent topics are undisclosed, making the topic discovery insufficient to understand documents \cite{dieng2020topic}.
        \item
            Topic collapsing damages the interpretability of topic models.
            It becomes difficult to infer the real underlying topics that a document contains \cite{Huynh2020}.
    \end{inparaenum}
    In consequence,
    topic collapsing impedes the utilization and extension of topic models;
    therefore it is crucial to overcome this challenge
    for building effective topic models.

    To address the topic collapsing issue and achieve effective topic modeling,
    we in this paper propose a novel neural topic model, \textbf{\modelfullname{}} (\textbf{\modelname{}}).
    First, 
    we illustrate the reason for topic collapsing:
    \Cref{fig_topic_collapse_ETM,fig_topic_collapse_NSTM,fig_topic_collapse_WeTe}
    show topic embeddings mostly collapse together in the semantic space of previous state-of-the-art methods.
    This makes discovered topics contain similar word semantics and thus results in the topic collapsing.
    Then to avoid the collapsing of topic embeddings,
    we propose the novel \textbf{\regfullname{}} (\textbf{\regname{}}) besides the reconstruction error of previous work.
    \regname{} regularizes topic embeddings as cluster centers and word embeddings as cluster samples.
    For effective regularization,
    \regname{} models the clustering soft-assignments between them by solving a specifically defined optimal transport problem on them.
    As such,
    \regname{} forces each topic embedding to be the center of a separately aggregated word embedding cluster.
    Instead of collapsing together, this makes topic embeddings away from each other and cover different semantics of word embeddings.
    Thus our ECR enables each produced topic to contain distinct word semantics, which alleviates topic collapsing. 
    Regularized by \regname{},
    our \modelname{}
    achieves effective topic modeling performance by producing diverse and coherent topics together with high-quality topic distributions of documents.
    \Cref{fig_topic_collapse_ECRTM} shows the effectiveness of \modelname{}.
    We conclude the main contributions of our paper as follows \footnote{Our code is available at \url{https://github.com/bobxwu/ECRTM}}:
    \begin{itemize}[leftmargin=*]
        \item
            We propose a novel embedding clustering regularization that avoids the collapsing of topic embeddings by forcing each topic embedding to be the center of a separately aggregated word embedding cluster,
            which effectively mitigates topic collapsing.
        \item
            We further propose a new neural topic model that jointly optimizes the topic modeling objective and the embedding clustering regularization objective.
            Our model can produce diverse and coherent topics with high-quality topic distributions of documents at the same time.
        \item
            We conduct extensive experiments on benchmark datasets and demonstrate that
            our model effectively addresses the topic collapsing issue and surpasses state-of-the-art baseline methods
            with substantially improved topic modeling performance.
    \end{itemize}

\section{Related Work} \label{sec_related_work}
    \paragraph{Conventional Topic Models}
        Conventional topic models \cite{hofmann1999probabilistic,blei2003latent,yao2014probabilistic,das2015gaussian}
        mostly employ probabilistic graphical models to model the generation of documents with topics as latent variables.
        They infer model parameters with MCMC methods like Gibbs sampling \cite{steyvers2007probabilistic} or Variational Inference \cite{blei2017variational}.
        Some studies use matrix factorization to model topics \cite{yan2013learning,kim2015simultaneous,shi2018short}.
        Many various scenarios have been developed, like short texts \cite{yan2013biterm,Wu2019}, multilingual \cite{mimno2009polylingual}, and dynamic topic models \cite{blei2006dynamic}.
        These methods commonly need model-specific derivations for different modeling assumptions.

    \paragraph{Neural Topic Models}
        Due to the success of Variational AutoEncoder \citep[VAE,][]{Kingma2014a,Rezende2014}, several neural topic models have been proposed \cite{Miao2016,Srivastava2017,dieng2019dynamic,meng2020discriminative,nguyen2021contrastive,Wu2021discovering,wu2022mitigating,wu2023infoctm}.
        Different from conventional ones,
        neural topic models can directly apply gradient back-propagation, which enhances flexibility and scalability.
        Alternatively,
        some studies directly cluster pre-trained word or sentence embeddings to produce topics \cite{Sia2020,zhang2022neural}, but they are \emph{not} topic models since they cannot infer the topic distributions of documents as required.
        Recent state-of-the-art NSTM \cite{zhao2020neural} and WeTe \cite{wang2022representing} measure the reconstruction error with optimal (conditional) transport distance.
        However,
        they still suffer from topic collapsing (see \Cref{sec_main_results}).
        Different from these,
        our proposed model aims to address the topic collapsing issue and
        achieve effective neural topic modeling.
        Besides the reconstruction error of these previous work,
        we propose a novel embedding clustering regularization that
        avoids the collapsing of topic embeddings by forcing each topic embedding to be the center of a separately aggregated word embedding cluster.
        Then our model learns topics under this effective regularization and particularly addresses the topic collapsing issue.

\begin{figure*}
    \centering
    \begin{subfigure}[t]{0.16666\linewidth}
        \centering
        \includegraphics[width=\linewidth]{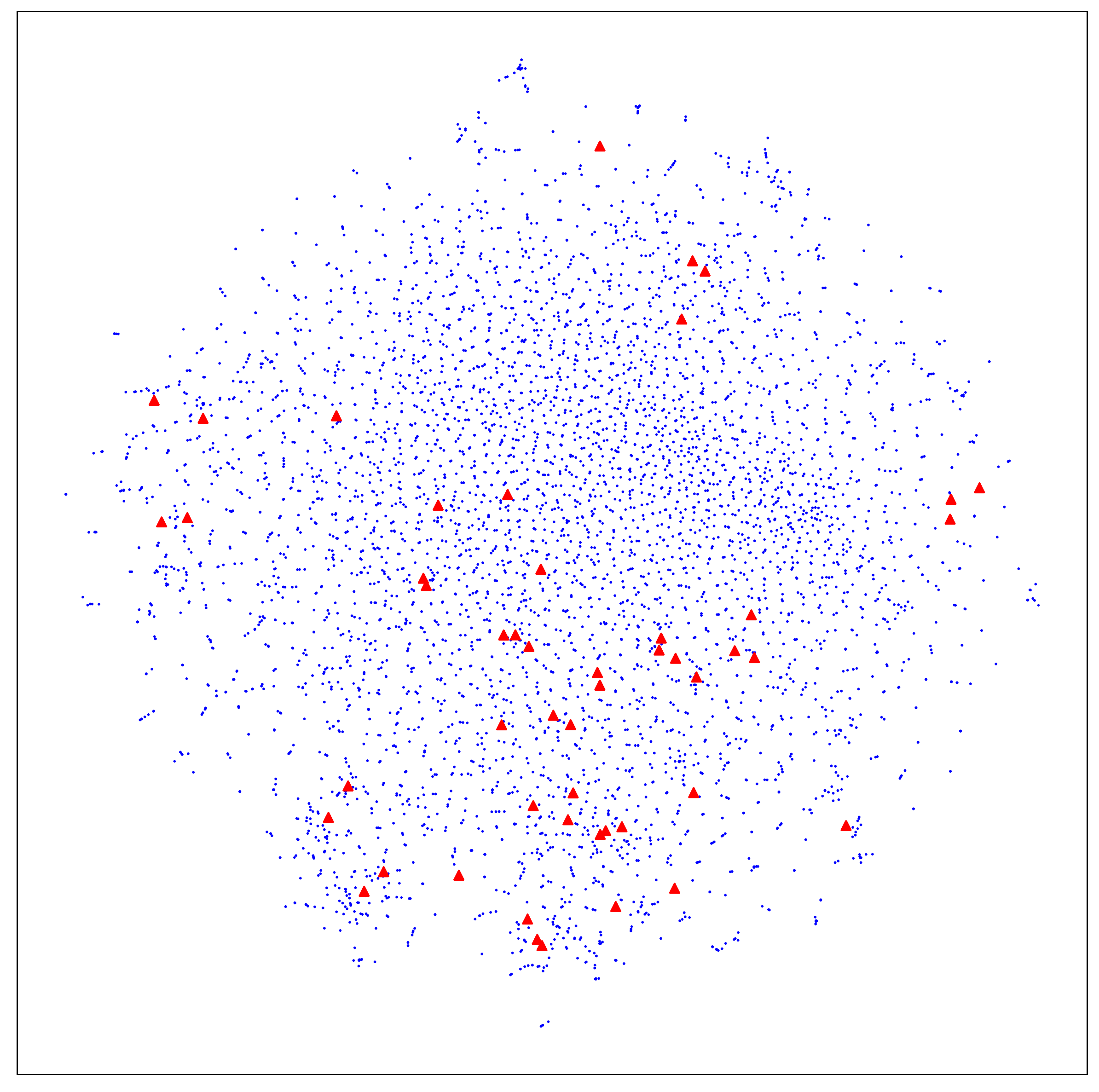}
        \caption{
            DKM
        }
        \label{fig_DKM}
    \end{subfigure}%
    \begin{subfigure}[t]{0.16666\linewidth}
        \centering
        \includegraphics[width=\linewidth]{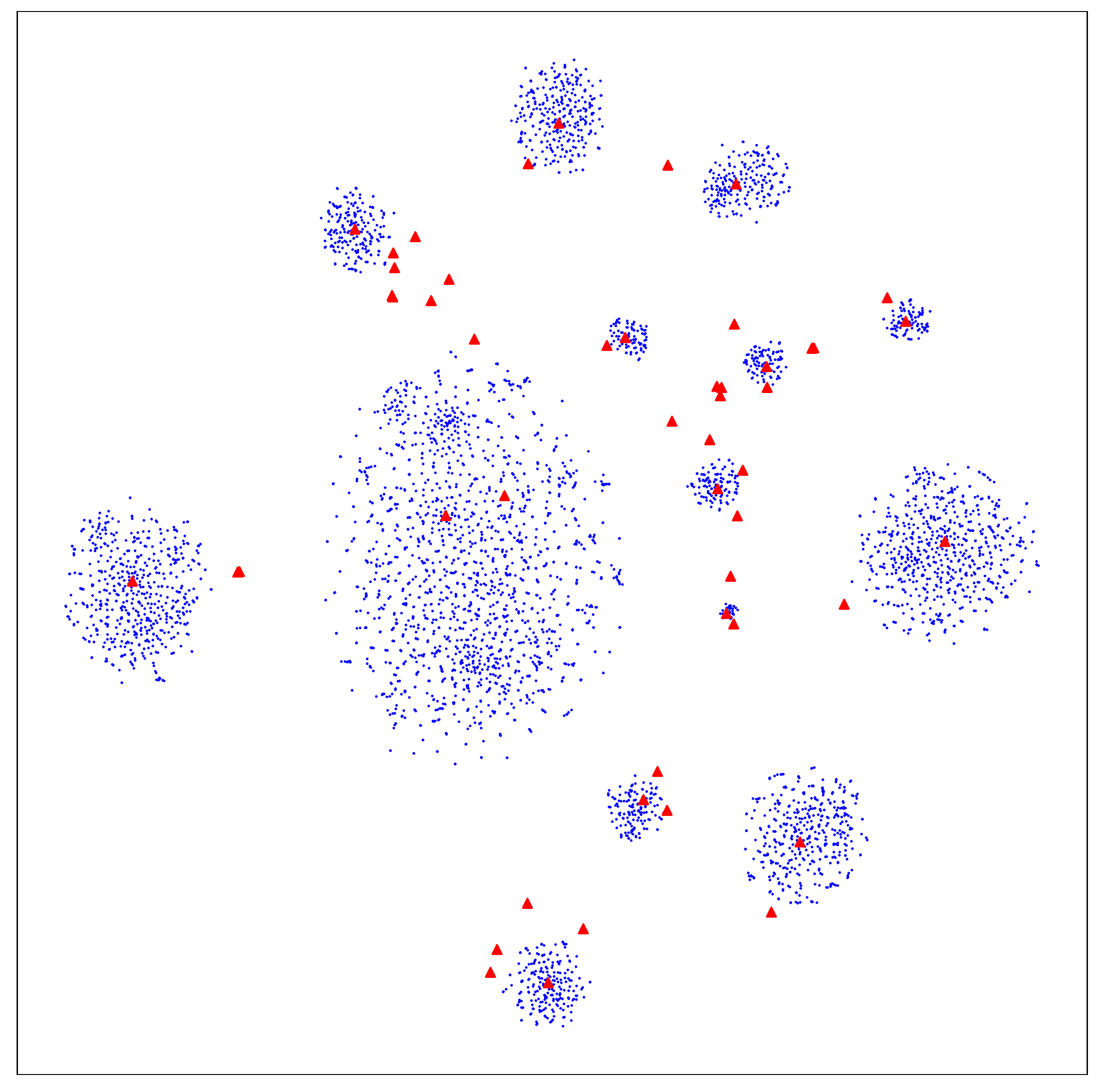}
        \caption{
            DKM+Entropy
        }
        \label{fig_DKM_entropy}
    \end{subfigure}%
    \begin{subfigure}[t]{0.16666\linewidth}
        \centering
        \includegraphics[width=\linewidth]{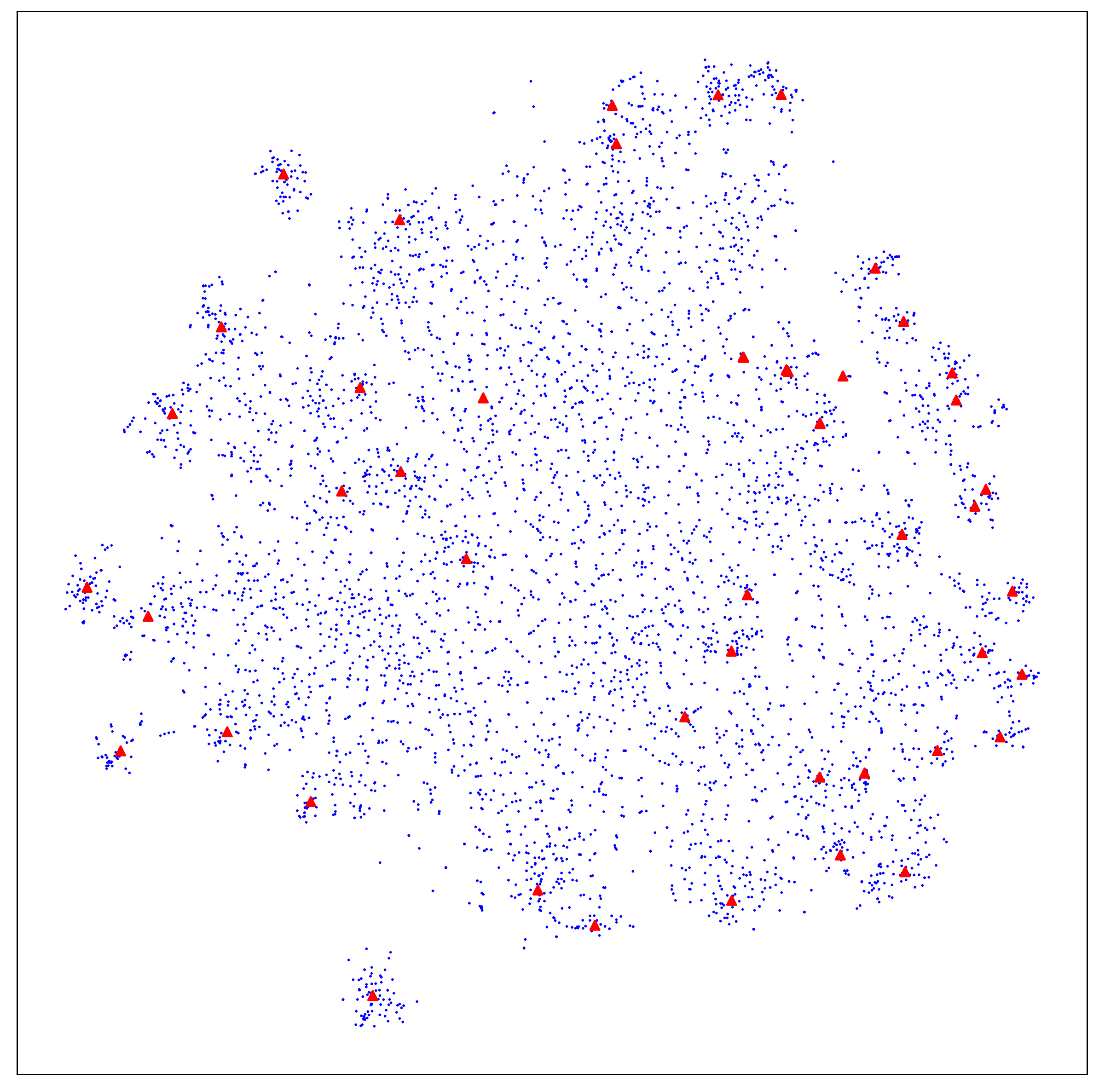}
        \caption{
            \textbf{\regname{}} ($\varepsilon \!\! = \!\! 1.0$)
        }
        \label{fig_epsilon_a}
    \end{subfigure}%
    \begin{subfigure}[t]{0.16666\linewidth}
        \centering
        \includegraphics[width=\linewidth]{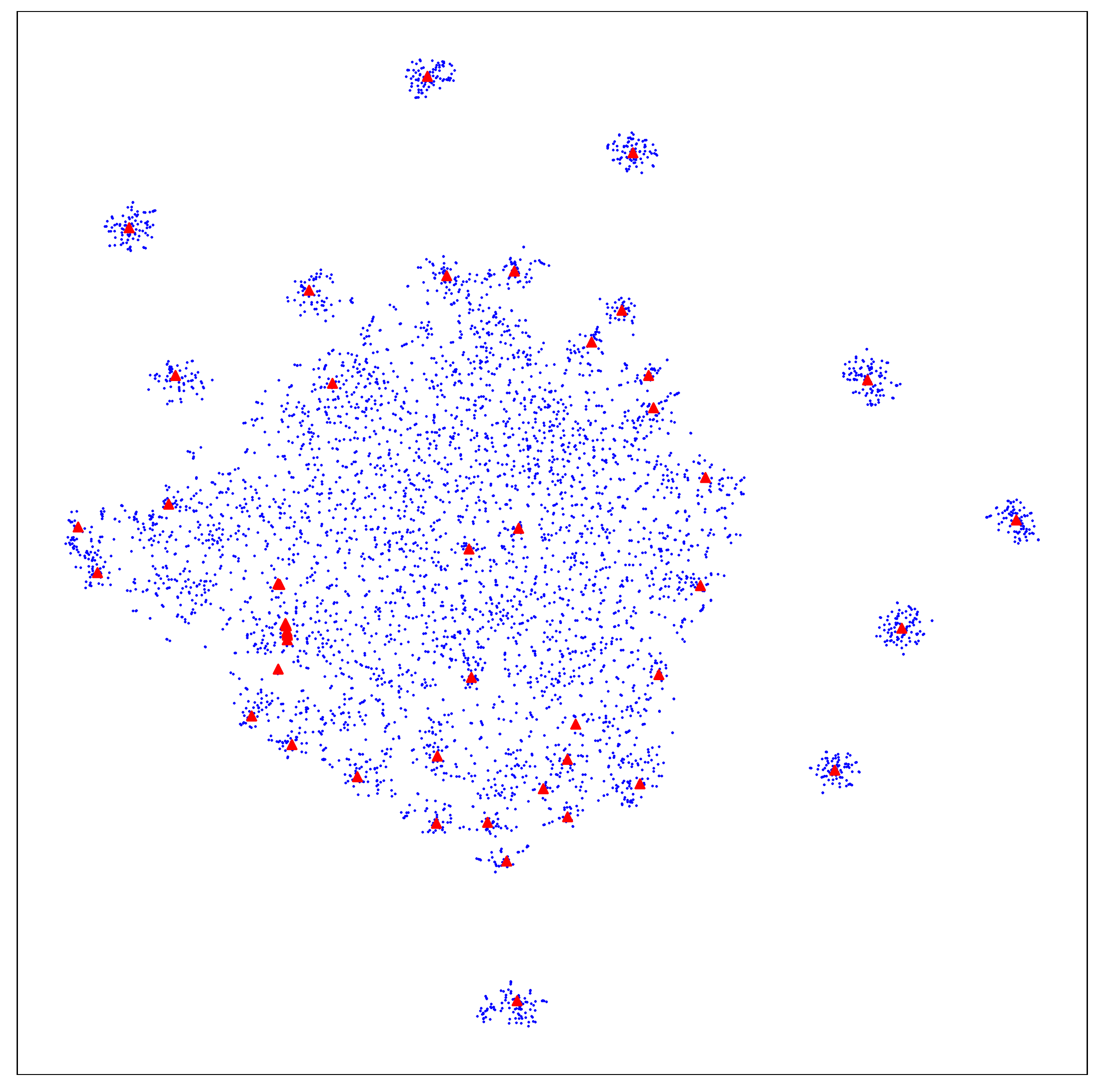}
        \caption{
            \textbf{\regname{}} ($\varepsilon \!\! = \!\! 0.1$)
        }
        \label{fig_epsilon_b}
    \end{subfigure}%
    \begin{subfigure}[t]{0.16666\linewidth}
        \centering
        \includegraphics[width=\linewidth]{input/img/tsne_ECRTM_K50}
        \caption{
            \textbf{\regname{}} ($\varepsilon \!\! = \!\! 0.05$)
        }
        \label{fig_sparse_assignment_ECRTM}
    \end{subfigure}%
    \begin{subfigure}[t]{0.16666\linewidth}
        \centering
        \includegraphics[width=\linewidth]{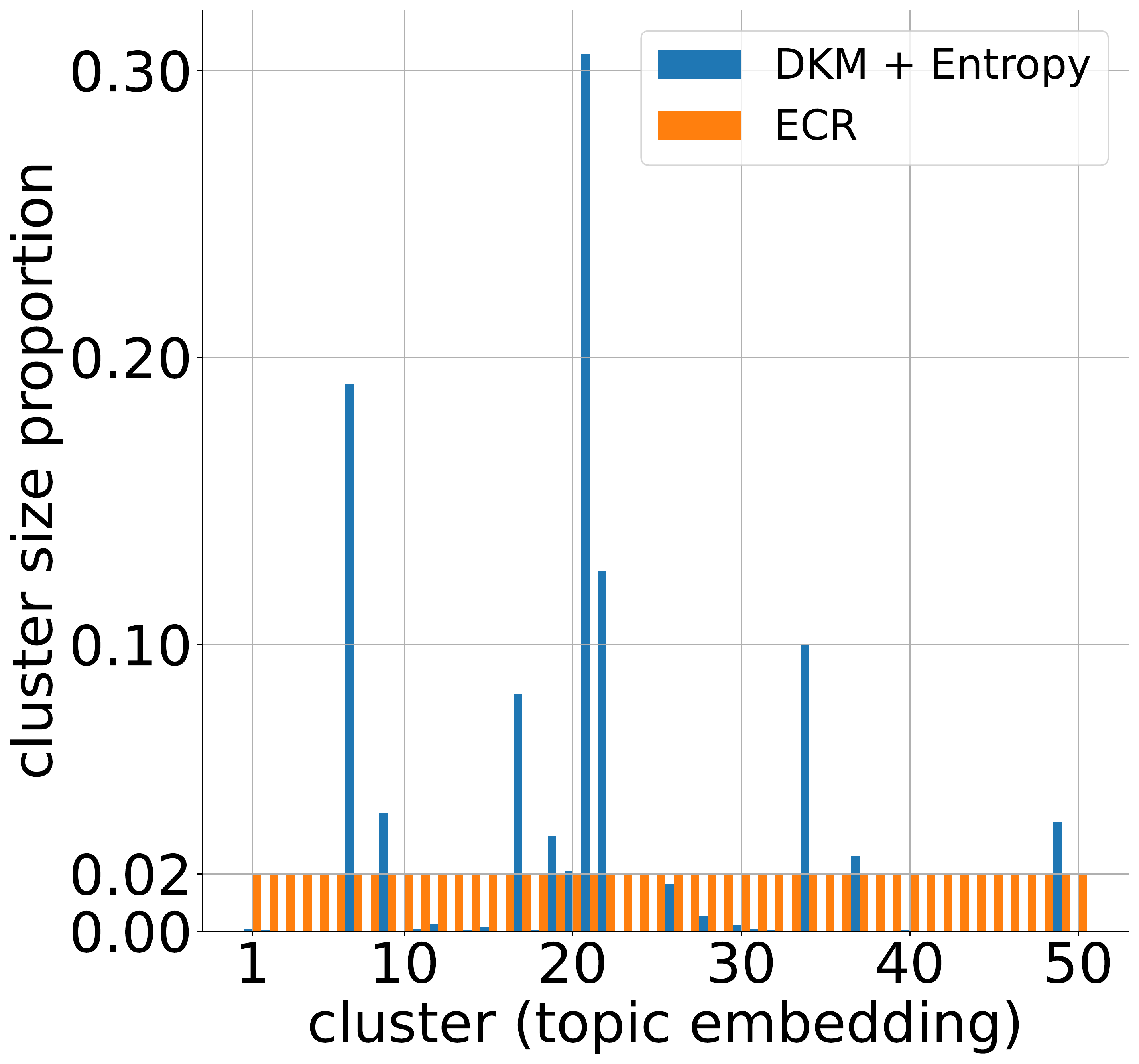}
        \caption{
            Cluster size proportions
        }
        \label{fig_cluster_size}
    \end{subfigure}%
    \caption{
        t-SNE visualization (a-e) of word embeddings ({\color{blue} $\bullet$}) and topic embeddings ({\color{red} $\blacktriangle$}) under 50 topics ($K\!\!\!=\!\!50$).
        (a): DKM cannot form separately aggregated clusters.
        (b): DKM + Entropy forms
        clusters but has a trivial solution that clusters of most topic embeddings are empty.
        (c,d,e):
        Our \regname{} forms
        clusters and also avoids the trivial solution of empty clusters with a small $\varepsilon$.
        (f):
        This quantitatively shows that while most cluster size proportions are zero in DKM + Entropy,
        our \regname{} successfully
        avoids this trivial solution
        with fulfilled preset cluster size constraints.
        Here we preset all cluster sizes as equal, so cluster size proportions all are $1/K\!\!=\!\!0.02$ (See \Cref{sec_reg}).
    }
    \label{fig_sparse_soft-assignment}
\end{figure*}

\section{Methodology}

    \subsection{Problem Setting and Notations}
        We recall the problem setting of topic modeling following LDA \cite{blei2003latent}.
        Consider a document collection $\mbf{X}$ with $V$ unique words (vocabulary size), and each document is denoted as $\mbf{x}$.
        We require to discover $K$ latent topics from this document collection.
        The $k$-th topic is defined as a distribution over all words (topic-word distribution), denoted as $\mbf{\beta}_{k} \!\! \in \!\! \mathbb{R}^{V}$.
        We have $\mbf{\beta} \!\! = \!\! (\mbf{\beta}_{1}, \dots, \mbf{\beta}_{K}) \!\! \in \!\! \mathbb{R}^{V \times K} $ as the topic-word distribution matrix of all topics.
        The topic distribution of a document (doc-topic distribution) refers to what topics it contains, denoted as $ \mbf{\theta} \! \in \! \Delta_{K} $.
        Here $\Delta_{K}$ denotes a probability simplex $\Delta_{K} \!\! = \!\! \left\{ \mbf{\theta} \in \mathbb{R}_{+}^{K} | \sum_{k=1}^{K} \! \theta_{k} \!\!=\!\! 1 \right\}$.

    \subsection{What Causes Topic Collapsing?} \label{sec_why}

        Despite the current achievements,
        most topic models suffer from \emph{topic collapsing}: topics semantically collapse towards each other (see \Cref{tab_motivation}).
        We illustrate what causes topic collapsing
        by analyzing a kind of recently proposed state-of-the-art neural topic models \cite{dieng2020topic,zhao2020neural}.
        These models
        compute the topic-word distribution matrix with two parameters: $\mbf{\beta} \!\!=\!\! \mbf{W}^{\top} \mbf{T}$.
        Here
        $\mbf{W} \!\! = \!\! ( \mbf{w}_1, \dots, \mbf{w}_{V} ) \!\! \in \!\! \mathbb{R}^{D \times V}$
        are the embeddings of $V$ words,
        and $\mbf{T} \!\! = \!\! ( \mbf{t}_1, \dots, \mbf{t}_{K} ) \!\! \in \!\! \mathbb{R}^{D \times K}$ are the embeddings of $K$ topics, all in the same $D$-dimensional semantic space.
        They can facilitate learning by initializing $\mbf{W}$ with pre-trained embeddings like
        GloVe \cite{pennington2014glove}.

        However, topic collapsing commonly happens in these state-of-the-art models.
        We believe the reason lies in
        that their reconstruction error minimization incurs the collapsing of topic embeddings.
        Specifically,
        these models learn topic and word embeddings by minimizing the reconstruction error between topic distribution $\mbf{\theta}$ and word distribution $\mbf{x}$ of a document.
        For example,
        to measure reconstruction error, ETM \cite{dieng2020topic} uses traditional expected log-likelihood, and recent NSTM \cite{zhao2020neural} and WeTe \cite{wang2022representing} use optimal (conditional) transport distance.
        In fact, words in a document collection commonly are long-tail distributed following Zipf's law \cite{reed2001pareto,piantadosi2014zipf}---roughly speaking, few words are of high frequency and most are of low frequency.
        Therefore the reconstruction is biased as it mainly reconstructs high-frequency words regardless of the reconstruction error measurements.
        Since topic and word embeddings are learned by minimizing reconstruction error,
        this biased reconstruction pushes most topic embeddings close to the embeddings of some high-frequency words in the semantic space.
        As a result, topic embeddings collapse together in these state-of-the-art methods as shown in \Cref{fig_motivation}.
        The topic-word distributions become similar to each other,
        leading to topic collapsing.
        We empirically demonstrate this argument by removing high-frequency words (See experiments in \Cref{sec_high-frequency}).

        \begin{figure*}[!t]
    \centering
    \begin{minipage}[b]{0.45\textwidth}
    \centering
    \begin{algorithm}[H]
    \footnotesize
    \caption{Training algorithm for \modelname{}.}
    \label{algo_training}
    \textbf{Input}: document collection $\mbf{X}$, preset cluster size constraint $\mbf{s}$, number of epochs $n_{\text{epoch}}$; \\
    \textbf{Output}: model parameters $\Theta$, $\mbf{W}$, $\mbf{T}$; \\
    \vspace{-\intextsep}
    \begin{algorithmic}[1]
        \FOR{$1$ to $n_{\text{epoch}}$}
            \FOR{each mini-batch $(\mbf{x}^{(1)}, \mbf{x}^{(2)}, \dots, \mbf{x}^{(N)})$ from $\mbf{X}$}

                \STATE \COMMENT{Sinkhorn's algorithm};
                \STATE $ C_{jk} = \| \mbf{w}_{j} - \mbf{t}_{k} \|^{2} \quad \forall \; j, k $;
                \STATE $ \mbf{M} = \exp(-\mbf{C} / \epsilon) $; %
                \STATE Initialize $ \mbf{b} \leftarrow \mbf{\mathds{1}}_{K} $;

                \WHILE{not converged and not reach max iterations}
                    \STATE $ \mbf{a} \leftarrow \frac{1}{V} \frac{\mbf{\mathds{1}}_{V}}{\mbf{M} \mbf{b}} $
                    , $ \mbf{b} \leftarrow \frac{\mbf{s}}{\mbf{M}^{\top} \mbf{a}}$;
                \ENDWHILE

                \STATE Compute $\mbf{\pi}_{\varepsilon}^{*} \leftarrow \mathrm{diag}(\mbf{a}) \mbf{M} \mathrm{diag}(\mbf{b})$;
                \STATE Compute $\mathcal{L}_{\text{TM}} + \lambda_{\mscript{\regname{}}} \mathcal{L}_{\mscript{\regname{}}}$ (\Cref{eq_full});
                \STATE Update $\Theta$, $\mbf{W}$, $\mbf{T}$ with a gradient step;
            \ENDFOR
        \ENDFOR
    \end{algorithmic}
\end{algorithm}
    \end{minipage}%
    \hfill
    \begin{minipage}[b]{0.53\textwidth}
        \centering
        \includegraphics[width=\linewidth]{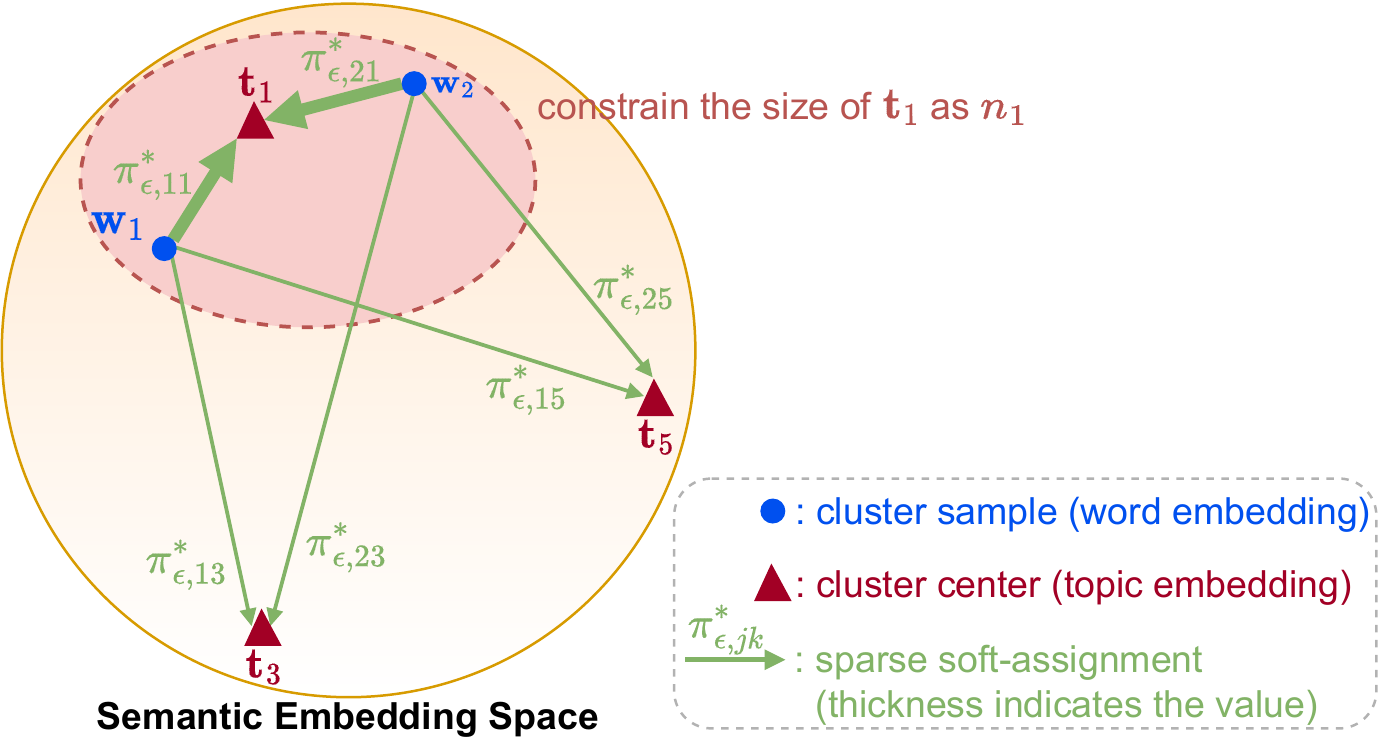}
        \caption{
            Illustration of \regname{}.
            \regname{} clusters word embeddings $\mbf{w}_j$ ({\color{blue} $\bullet$}) as samples and topic embeddings $\mbf{t}_k$ ({\color{red} $\blacktriangle$}) as centers
            with soft-assignment $\pi_{\epsilon, jk}^{*}$.
            The cluster size of center $\mbf{t}_k$ is constrained as $n_k$.
            Here \regname{} pushes $\mbf{w}_1$ and $\mbf{w}_2$ close to $\mbf{t}_1$ and away from $\mbf{t}_3$ and $\mbf{t}_5$.
        }
    \label{fig_illustration}
    \end{minipage}%
    
    \vspace{-\intextsep}

\end{figure*}

        \subsection{How to Design An Effective Regularization on Embeddings?} \label{sec_how_to}
            In this section, we discuss how to design an effective regularization on embeddings for the topic collapsing issue.

            Our analysis in \Cref{sec_why} indicates topic collapsing happens
            because the reconstruction error minimization incurs the collapsing of topic embeddings in existing work.
            To address this issue,
            we propose to design a clustering regularization on embeddings
            in addition to the reconstruction error of existing work.
            We consider topic embeddings as cluster centers and word embeddings as cluster samples;
            then we require the regularization to
            force each topic embedding to be the center of a separately aggregated word embedding cluster.
            As such, instead of collapsing together, topic embeddings are away from each other and cover different semantics of word embeddings in the space.
            This will make each discovered topic contain distinct word semantics and thus alleviate topic collapsing.
            \textbf{However, it is \textit{non-trivial} to design such an effective regularization.}
            We explore the requirements as follows.

            \paragraph{Supporting Joint Optimization}
                As we regularize on neural topic models,
                we require the clustering regularization to support joint optimization
                on topic and word embeddings along with a neural topic modeling objective.
                Some studies \cite{Sia2020} apply classic clustering algorithms, \emph{e.g.}, KMeans and GMM,
                to produce topics by clustering pre-trained word embeddings.
                We clarify that they are \textit{not} topic models as they only produce topics and cannot learn doc-topic distributions as required (but we compare them in experiments).
                We do not adopt these classic clustering algorithms and some other work \cite{song2013auto,huang2014deep,xie2016unsupervised,hsu2017cnn,yang2017towards} as our clustering regularization,
                because we cannot jointly optimize them along with a neural topic modeling objective.

            \paragraph{Producing Sparse Soft-assignments}
                We also require the clustering regularization to produce sparse soft-assignments.
                Even supporting joint optimization, existing clustering methods may still lead to topic collapsing.
                For example,
                we propose to use the state-of-the-art deep clustering method, Deep KMeans \citep[DKM,][]{fard2020deep} that supports joint optimization.
                \textbf{Note we are the first to use DKM in topic modeling.}
                Its clustering objective 
                is to minimize the total Euclidean distance between centers and samples weighted by soft-assignments.
                We use DKM as a clustering regularization on topic and word embeddings:
                \begin{align}
                     &\min_{\mbf{W}, \mbf{T}, \mbf{p}}  \sum_{j=1}^{V} \sum_{k=1}^{K} \| \mbf{w}_{j} - \mbf{t}_{k} \|^{2} p_{jk}, \notag \\
                     &\text{where} \quad
                    p_{jk} = \frac{e^{ - \| \mbf{w}_{j} - \mbf{t}_{k} \|^{2} / \tau}}{ \sum_{k'=1}^{K} e^{ - \| \mbf{w}_{j} - \mbf{t}_{k'} \|^{2} / \tau} }.
                    \label{eq_DKM}
                \end{align}
                Here $p_{jk}$ denotes the clustering soft-assignment of word embedding $\mbf{w}_{j}$ assigned to topic embedding $\mbf{t}_{k}$, which is modeled as a 
                softmax function of the Euclidean distance between $\mbf{w}_{j}$ and all topic embeddings ($\tau$ is a temperature hyperparameter).
                Unfortunately,
                DKM still incurs topic collapsing (See quantitative results in \Cref{sec_ablation}).
                We see from \Cref{fig_DKM} that DKM cannot form separately aggregated clusters, so the topic embeddings (centers)
                cannot be separated but collapse together.
                To solve this issue,
                we require the clustering regularization to produce sparse soft-assignments---each word embedding is mainly assigned to only one topic embedding and rarely to others, which pushes each word embedding only close to one topic embedding and away from all others in the semantic space.
                This way can form separately aggregated word embedding clusters with topic embeddings as centers, which encourages topic embeddings to be away from each other.
                \textbf{Note that we do \emph{not} directly model latent topics with these sparse soft-assignments (See \Cref{sec_ECRTM}).} 

            \paragraph{Fulfilling Preset Cluster Size Constraints}
                We further require the clustering regularization to fulfill preset cluster size constraints.
                Only producing sparse soft-assignments may still result in topic collapsing.
                To make the soft-assignments sparse,
                we propose DKM+Entropy that jointly minimizes the entropy of soft-assignments, $\sum_{j=1}^{V} \sum_{k=1}^{K} -p_{jk} \log p_{jk}$, with the clustering objective of DKM~(\Cref{eq_DKM}).
                However, this way still leads to topic collapsing (See quantitative results in \Cref{sec_ablation}).
                \Cref{fig_DKM_entropy} shows
                DKM+Entropy indeed forms separately aggregated clusters for some topic embeddings,
                but unfortunately the clustering solution is trivial---the clusters of most topic embeddings are empty,
                as quantitatively shown in \Cref{fig_cluster_size}.
                As a result,
                the topic embeddings of these empty clusters
                cannot be separated to cover distinct semantics but only collapse to others in the space.
                To avoid such trivial solutions of empty clusters,
                we propose to preset constraints on the size of each cluster (must not be empty)
                and require the clustering regularization to fulfill these constraints.

        \subsection{Embedding Clustering Regularization} \label{sec_reg}
            To meet the above necessary requirements,
            we in this section introduce a novel method, \textbf{\regfullname{}} (\textbf{\regname{}}).
            \Cref{fig_illustration} illustrates \regname{}, and \Cref{fig_sparse_assignment_ECRTM} shows its effectiveness.

            \paragraph{Presetting Cluster Size Constraints}
            We first preset the cluster size constraints to be fulfilled to avoid trivial solutions of empty clusters.
            Recall that we have $K$ topic embeddings as centers and $V$ word embeddings as samples.
            We denote the cluster size of topic embedding $\mbf{t}_k$ as $n_k$
            and the cluster size proportion as $s_{k} \!\!=\!\! n_{k}/V$.
            We have $\mbf{s} \!=\! (s_{1}, \dots, s_{K}) ^{\top} \!\! \in \!\! \Delta_{K}$ as the vector of all cluster size proportions.
            Unfortunately,
            we usually lack prior knowledge about the cluster sizes of topic embeddings.
            Previous studies \cite{wallach2009rethinking} find that a symmetric Dirichlet prior over topic-word distributions achieves better performance in LDA.
            Inspired by this,
            we set all cluster sizes as uniform: $n_k \!\!=\!\! V/K$ and $\mbf{s} \!\!=\!\! (1/K, \dots, 1/K)^{\top}$.
            This setting can avoid the trivial solutions of empty clusters, and experiments show it works well across datasets (see \Cref{sec_main_results}).
            Note that $\mbf{s}$ can be different values determined by prior knowledge from experts, and we leave this as future work.

            \paragraph{Embedding Clustering Regularization (ECR)}
            To meet the above requirements,
            we propose ECR that
            models clustering soft-assignments with the transport plan of a specifically defined optimal transport problem.
            Specifically,
            we define two discrete measures of topic ($\mbf{t}_k$) and word embeddings ($\mbf{w}_j$): $\gamma \!\!=\!\! \sum_{j=1}^{V} \frac{1}{V} \delta_{\mbf{w}_{j}}$ and $ \phi \!\!=\!\! \sum_{k=1}^{K} s_{k} \delta_{\mbf{t}_{k}} $, where $\delta_{x}$ denotes the Dirac unit mass on $x$.
            We formulate the entropic regularized optimal transport between $\gamma$ and $\phi$ as
            \begin{align}
                & \argmin_{\mbf{\pi} \in \mathbb{R}_{+}^{V \times K}} \mathcal{L}_{\mscript{OT}_{\varepsilon}} (\gamma, \phi), \quad \mathcal{L}_{\mscript{OT}_{\varepsilon}} (\gamma, \phi) = \notag \\
                &  \sum_{j=1}^{V} \! \sum_{k=1}^{K} \! \| \mbf{w}_{j} \!-\! \mbf{t}_{k} \|^{2} \pi_{jk} + \sum_{j=1}^{V} \! \sum_{k=1}^{K} \varepsilon \pi_{jk} (\log (\pi_{jk}) \!-\! 1 ) \notag \\
                & \text{s.t.} \quad \mbf{\pi} \mbf{\mathds{1}}_{K} = \frac{1}{V} \mbf{\mathds{1}}_{V} \quad \text{and} \quad \mbf{\pi}^{\top} \mbf{\mathds{1}}_{V} = \mbf{s}. \label{eq_OT_regularized}
            \end{align}
            Here the first term of $\mathcal{L}_{\mscript{OT}_{\varepsilon}}$ is the original optimal transport problem, and the second term with hyperparameter $\varepsilon$ is the entropic regularization to make this problem tractable \cite{canas2012learning}.
            \Cref{eq_OT_regularized} is to find the optimal transport plan $\mbf{\pi}_{\varepsilon}^{*}$ that minimizes
            the total cost of transporting weight from word embeddings to topic embeddings.
            We measure the transport cost between word embedding $\mbf{w}_j$ and topic embedding $\mbf{t}_k$ by Euclidean distance:
            $C_{jk} \!\!=\!\! \|\mbf{w}_j - \mbf{t}_k \|^2$, and the transport cost matrix is $\mbf{C} \!\! \in \!\! \mathbb{R}^{V \times K}$.
            The two conditions in \Cref{eq_OT_regularized} restrict the weight of each word embedding $\mbf{w}_j$ as $\frac{1}{V}$ and each topic embedding $\mbf{t}_k$ as $s_k$,
            where $\mbf{\mathds{1}}_{K}$ ($\mbf{\mathds{1}}_{V}$) is a $K$ ($V$) dimensional column vector of ones.
            ${\pi}_{jk}$ denotes the transport weight from $\mbf{w}_j$ to $\mbf{t}_k$;
            $\mbf{\pi} \!\! \in \!\! \mathbb{R}_{+}^{V \times K} $ is the transport plan
            that includes the transport weight of each word embedding to fulfill the weight of each topic embedding.

            To meet the above requirements,
            we model clustering soft-assignments with the optimal transport plan $\mbf{\pi}_{\varepsilon}^{*}$, \emph{i.e.},
            the soft-assignment of $\mbf{w}_j$ to $\mbf{t}_k$ is the transport weight between them, $\pi^{*}_{\varepsilon, jk}$.
            Then we formulate our \regname{} objective by minimizing the total distance between word and topic embeddings weighted by these soft-assignments:
            \begin{align}
                &\mathcal{L}_{\mscript{\regname{}}} = \sum_{j=1}^{V} \sum_{k=1}^{K} \| \mbf{w}_{j} - \mbf{t}_{k} \|^{2} \pi^{*}_{\varepsilon, jk}, \quad\quad \notag \\
                &
                \text{where} \,\,
                \mbf{\pi}_{\varepsilon}^{*} \!=\! \mathrm{sinkhorn} (\gamma, \phi, \varepsilon)
                \! \approx \! \argmin_{\mbf{\pi} \in \mathbb{R}_{+}^{V \times K}} \mathcal{L}_{\mscript{OT}_{\varepsilon}} (\gamma, \phi).
                \label{eq_clustering_regularization}
            \end{align}
            To solve this specifically defined optimal transport problem,
            here we compute
            $\mbf{\pi}_{\varepsilon}^{*}$ through Sinkhorn's algorithm \citep{sinkhorn1964relationship,cuturi2013lightspeed}, a fast iterative scheme 
            suited to the execution of GPU \cite{peyre2019computational}.
            See \Cref{algo_training} for detailed algorithm steps.
            By doing so, $\mbf{\pi}_{\varepsilon}^{*}$ is a differentiable variable
            parameterized by transport cost matrix $\mbf{C}$ \cite{salimans2018improving,genevay2018learning}.
            Due to this,
            minimizing $\pi^{*}_{\varepsilon, jk}$ increases transport cost $C_{jk}$, \emph{i.e.}, the distance between $\mbf{w}_j$ and $\mbf{t}_k$; otherwise maximizing it decreases the distance \cite{Genevay2019Different}.
            Thus we can exactly model $\mbf{\pi}_{\varepsilon}^{*}$ as differentiable clustering soft-assignments between topic and word embeddings.

            \paragraph{ECR is an Effective Regularization on Embeddings}
            First,
            \regname{} supports joint optimization
            since $\mbf{\pi}^{*}_{\varepsilon}$ is differentiable as aforementioned.
            Second,
            \regname{} produces sparse soft-assignments.
            It is proven that $\mbf{\pi}^{*}_{\varepsilon}$ converges to the optimal solution of the original optimal transport problem when $\varepsilon \! \rightarrow \! 0 $, which leads to a sparse transport plan \cite{peyre2019computational}.
            Hence \regname{} (\Cref{eq_clustering_regularization}) produces sparse soft-assignments
            under a small $\varepsilon$.
            With sparse soft-assignments,
            \regname{}
            pushes each word embedding only close to one topic embedding and away from all others, which forms separately aggregated clusters.
            We illustrate this property in \Cref{fig_epsilon_a,fig_epsilon_b,fig_sparse_assignment_ECRTM}.
            Last,
            \regname{} fulfills preset cluster size constraints.
            In \Cref{eq_OT_regularized},
            the transport plan is restricted by two conditions
            indicating the weight of each word embedding $\mbf{w}_j$ is $\frac{1}{V}$
            and each topic embedding $\mbf{t}_k$ is $s_k$.
            These ensure the sparse optimal transport plan $\mbf{\pi}^{*}_{\varepsilon}$ needs to transport $n_k$ word embeddings to topic embedding $\mbf{t}_k$ to balance the weight, such that $n_k \!\! \times \!\! \frac{1}{V} \!\!=\!\! s_k$.
            Accordingly,
            \regname{} fulfills the preset cluster size constraints with $\mbf{\pi}^{*}_{\varepsilon}$ as clustering soft-assignments.
            This avoids trivial clustering solutions of empty clusters as shown in \Cref{fig_cluster_size}.

            To sum up, our \regname{} effectively forces each topic embedding to be the center of a separately aggregated word embedding cluster.
            This makes topic embeddings away from each other and cover different semantics of word embeddings,
            which alleviates topic collapsing.

    \subsection{\modelfullname{}} \label{sec_ECRTM}
        In this section, 
        we propose a novel topic model, \textbf{\modelfullname} (\textbf{\modelname{})}
        that jointly optimizes the topic modeling objective
        and the \regname{} objective.
        \Cref{algo_training} shows its training algorithm.

        \paragraph{Inferring Doc-Topic Distributions}
            We devise the prior and variational distribution following VAE \cite{Kingma2014a} to infer doc-topic distributions.
            In detail, we draw a latent variable, $\mbf{r}$, from a
            logistic normal distribution:
            $ p(\mbf{r}) \!=\! \mathcal{LN} (\mbf{\mu}_{0}, \mbf{\Sigma}_{0}) $,
            where $\mbf{\mu}_{0}$ and $\mbf{\Sigma}_{0}$ are the mean and diagonal covariance matrix \cite{Srivastava2017}.
            Then we use an encoder network that outputs parameters of the variational distribution, the mean vector $ \mbf{\mu} \!=\! f_{\mbf{\mu}}(\mbf{x}) $ and covariance matrix $ \mbf{\Sigma} \!=\! \mathrm{diag}( f_{\mbf{\Sigma}}(\mbf{x}) ) $.
            So the variational distribution is $ q_{\Theta}(\mbf{r} | \mbf{x}) \!=\! \mathcal{N}(\mbf{\mu}, \mbf{\Sigma})$ where $\Theta$ denotes the parameters of $f_{\mbf{\mu}}$ and $f_{\mbf{\Sigma}}$.
            By applying the reparameterization trick \cite{Kingma2014a}, we sample
            $\mbf{r} \!=\! \mbf{\mu} + \mbf{\Sigma}^{1/2} \mbf{\epsilon}$ where $\mbf{\epsilon} \sim \mathcal{N}(0, \mbf{I})$.
            We obtain the doc-topic distribution $\mbf{\theta}$ with a softmax function as $\mbf{\theta} \!=\! \mathrm{softmax} ( \mbf{r} )$.

        \paragraph{Reconstructing Documents}
            We then reconstruct the input documents with topic-word distribution matrix $\mbf{\beta} \in \mathbb{R}^{V \times K} $.
            Recall that $\mbf{\beta}$ indicates the weights between all topics and words.
            Previous methods commonly model $\mbf{\beta}$ as the product of topic and word embeddings (\Cref{sec_why}).
            Differently,
            our model uses the proposed ECR as a clustering regularization on topic and word embeddings,
            so $\mbf{\beta}$ also needs to reflect the learned clustering assignments between them.
            \textbf{We do \textit{not} directly model $\mbf{\beta}$ with the soft-assignments $\mbf{\pi}^{*}_{\varepsilon}$ of our \regname{}.}
            This is because it makes one word only belong to one topic as $\mbf{\pi}^{*}_{\varepsilon}$ is very sparse (most values are close to zero as aforementioned);
            but in reality, one word should be able to belong to different topics \cite{blei2003latent}.
            To this end, we propose to model $\mbf{\beta}$ as
            \begin{equation}
                \beta_{jk} = \frac{e^{ - \| \mbf{w}_{j} - \mbf{t}_{k} \|^{2} / \tau}}{ \sum_{k'=1}^{K} e^{ - \| \mbf{w}_{j} - \mbf{t}_{k'} \|^{2} / \tau} } \label{eq_beta}
            \end{equation}
            where $\tau$ is a temperature hyperparameter. %
            This formulation is similar to the less sparse soft-assignments of DKM \cite{fard2020deep}.
            It is less sparse and can reflect the learned clustering assignments between topic and word embeddings.
            Thus one word can belong to different topics, which agrees with reality.
            With the doc-topic distribution $\mbf{\theta}$ and the topic-word distribution matrix $\mbf{\beta}$, we routinely sample the reconstructed document from a Multinomial distribution $\mathrm{Multi}(\mathrm{softmax}(\mbf{\beta} \mbf{\theta}))$.

        \paragraph{Overall Objective Function of ECRTM}
            Given a batch of $N$ documents $( \mbf{x}^{(1)}, \dots, \mbf{x}^{(N)} )$, we write the topic modeling objective function following VAE as
            \begin{align}
                \mathcal{L}_{\mscript{TM}} &= \frac{1}{N} \sum_{i=1}^{N} -(\mbf{x}^{(i)})^{\top} \log (\mathrm{softmax}(\mbf{\beta} \mbf{\theta}^{(i)})) \notag \\
                &+ \mathrm{KL} \left[ q_{\Theta}(\mbf{r}^{(i)} | \mbf{x}^{(i)}) \| p(\mbf{r}^{(i)}) \right]. \label{eq_TM}
            \end{align}
            The first term is the reconstruction error, and the second term is the KL divergence between the prior and variational distribution.
            \modelname{} learns topics regularized by our \regname{}.
            We define the overall objective function of \modelname{} as a combination of $\mathcal{L}_{\text{TM}}$ (\Cref{eq_TM}) and $\mathcal{L}_{\mscript{\regname{}}}$ (\Cref{eq_clustering_regularization}):
            \begin{equation}
                \min_{\Theta, \mbf{W}, \mbf{T}} \mathcal{L}_{\mscript{TM}} + \lambda_{\mscript{\regname{}}} \; \mathcal{L}_{\mscript{\regname{}}} \label{eq_full}
            \end{equation}
            where $\lambda_{\mscript{\regname{}}}$ is a weight hyperparameter.
            This overall objective enables \modelname{} to aggregate the embeddings of related words to form separate clusters with topic embeddings as centers and avoids the collapsing of topic embeddings.
            Thus our \modelname{} can alleviate the topic collapsing issue and learn coherent and diverse topics together with high-quality doc-topic distributions at the same time.

\begin{table*}[!ht]
    \centering
    \setlength{\tabcolsep}{1.8mm}
    \renewcommand{\arraystretch}{1.2}
    \resizebox{\linewidth}{!}{
    \begin{tabular}{lrrrrrrrrrrrrrrrrrrr}
    \toprule
    \multicolumn{1}{c}{\multirow{3}[4]{*}{Model}} & \multicolumn{4}{c}{20NG}      &       & \multicolumn{4}{c}{IMDB}      &       & \multicolumn{4}{c}{Yahoo Answer} &       & \multicolumn{4}{c}{AG News} \\
    \cmidrule{2-5}\cmidrule{7-10}\cmidrule{12-15}\cmidrule{17-20}          & \multicolumn{2}{c}{$K$=50} & \multicolumn{2}{c}{$K$=100} &       & \multicolumn{2}{c}{$K$=50} & \multicolumn{2}{c}{$K$=100} &       & \multicolumn{2}{c}{$K$=50} & \multicolumn{2}{c}{$K$=100} &       & \multicolumn{2}{c}{$K$=50} & \multicolumn{2}{c}{$K$=100} \\
          & \multicolumn{1}{c}{$C_V$} & \multicolumn{1}{c}{TD} & \multicolumn{1}{c}{$C_V$} & \multicolumn{1}{c}{TD} &       & \multicolumn{1}{c}{$C_V$} & \multicolumn{1}{c}{TD} & \multicolumn{1}{c}{$C_V$} & \multicolumn{1}{c}{TD} &       & \multicolumn{1}{c}{$C_V$} & \multicolumn{1}{c}{TD} & \multicolumn{1}{c}{$C_V$} & \multicolumn{1}{c}{TD} &       & \multicolumn{1}{c}{$C_V$} & \multicolumn{1}{c}{TD} & \multicolumn{1}{c}{$C_V$} & \multicolumn{1}{c}{TD} \\
    \midrule
    LDA   & $^\ddag$0.385 & $^\ddag$0.655 & $^\ddag$0.387 & $^\ddag$0.622 &       & $^\ddag$0.347 & $^\ddag$0.788 & $^\ddag$0.342 & $^\ddag$0.691 &       & $^\ddag$0.359 & $^\ddag$0.843 & $^\ddag$0.359 & $^\ddag$0.602 &       & $^\ddag$0.364 & $^\ddag$0.864 & $^\ddag$0.349 & $^\ddag$0.696 \\
    KM    & $^\ddag$0.251 & $^\ddag$0.204 & $^\ddag$0.294 & $^\ddag$0.317 &       & $^\ddag$0.213 & $^\ddag$0.219 & $^\ddag$0.244 & $^\ddag$0.302 &       & $^\ddag$0.271 & $^\ddag$0.242 & $^\ddag$0.297 & $^\ddag$0.345 &       & $^\ddag$0.241 & $^\ddag$0.264 & $^\ddag$0.289 & $^\ddag$0.395 \\
    WLDA  & $^\ddag$0.378 & $^\ddag$0.375 & $^\ddag$0.369 & $^\ddag$0.273 &       & $^\ddag$0.311 & $^\ddag$0.053 & $^\ddag$0.320 & $^\ddag$0.069 &       & $^\ddag$0.333 & $^\ddag$0.322 & $^\ddag$0.338 & $^\ddag$0.292 &       & $^\ddag$0.384 & $^\ddag$0.410 & $^\ddag$0.378 & $^\ddag$0.323 \\
    DVAE  & $^\ddag$0.331 & $^\ddag$0.598 & $^\ddag$0.372 & $^\ddag$0.658 &       & $^\ddag$0.294 & $^\ddag$0.050 & $^\ddag$0.290 & $^\ddag$0.229 &       & $^\ddag$0.338 & $^\ddag$0.674 & $^\ddag$0.376 & $^\ddag$0.589 &       & $^\ddag$0.419 & $^\ddag$0.347 & $^\ddag$0.298 & $^\ddag$0.113 \\
    ETM   & $^\ddag$0.375 & $^\ddag$0.704 & $^\ddag$0.369 & $^\ddag$0.573 &       & $^\ddag$0.346 & $^\ddag$0.557 & $^\ddag$0.341 & $^\ddag$0.371 &       & $^\ddag$0.354 & $^\ddag$0.719 & $^\ddag$0.353 & $^\ddag$0.624 &       & $^\ddag$0.364 & $^\ddag$0.819 & $^\ddag$0.371 & $^\ddag$0.773 \\
    HyperMiner & $^\ddag$0.371 & $^\ddag$0.613 & $^\ddag$0.368  & $^\ddag$0.446  &       & $^\ddag$0.347  & $^\ddag$0.485  & $^\ddag$0.343  & $^\ddag$0.258 &       & $^\ddag$0.344 & $^\ddag$0.507 & $^\ddag$0.346 & $^\ddag$0.444 &       & $^\ddag$0.359 & $^\ddag$0.521 & $^\ddag$0.360 & $^\ddag$0.343 \\
    NSTM  & $^\ddag$0.395 & $^\ddag$0.427 & $^\ddag$0.391 & $^\ddag$0.473 &       & $^\ddag$0.334 & $^\ddag$0.175 & $^\ddag$0.340 & $^\ddag$0.255 &       & $^\ddag$0.390 & $^\ddag$0.658 & 0.387 & $^\ddag$0.659 &       & $^\ddag$0.411 & $^\ddag$0.873 & \textbf{0.421} & $^\ddag$0.832 \\
    WeTe  & $^\ddag$0.383 & $^\ddag$0.949 & $^\ddag$0.352 & $^\ddag$0.742 &       & $^\ddag$0.368 & $^\ddag$0.931 & $^\ddag$0.293 & $^\ddag$0.638 &       & $^\ddag$0.367 & $^\ddag$0.878 & $^\ddag$0.353 & $^\ddag$0.544 &       & $^\ddag$0.383 & $^\ddag$0.945 & $^\ddag$0.363 & $^\ddag$0.827 \\
    \midrule
    \textbf{\modelname{}} & \textbf{0.431} & \textbf{0.964} & \textbf{0.405} & \textbf{0.904} &       & \textbf{0.393} & \textbf{0.974} & \textbf{0.373} & \textbf{0.887} &       & \textbf{0.405} & \textbf{0.985} & \textbf{0.389} & \textbf{0.903} &       & \textbf{0.466} & \textbf{0.961} & 0.416 & \textbf{0.981} \\
    \bottomrule
    \end{tabular}%
    }
    \caption{
        Topic quality of topic coherence ($C_V$) and topic diversity (TD) under 50 and 100 topics ($K\!\!=\!\!50$ and $K\!\!=\!\!100$). The best scores are in \textbf{bold}.
        $\ddag$ means the gain of \modelname{} is statistically significant at 0.05 level.
    }
    \label{tab_topic_quality}%
\end{table*}%

\begin{table*}[!ht]
    \centering
    \setlength{\tabcolsep}{1.8mm}
    \renewcommand{\arraystretch}{1.2}
    \resizebox{\linewidth}{!}{
    \begin{tabular}{lrrrrrrrrrrrrrrrrrrr}
      \toprule
      \multicolumn{1}{c}{\multirow{3}[4]{*}{Model}} & \multicolumn{4}{c}{20NG}      &       & \multicolumn{4}{c}{IMDB}      &       & \multicolumn{4}{c}{Yahoo Answer} &       & \multicolumn{4}{c}{AG News} \\
  \cmidrule{2-5}\cmidrule{7-10}\cmidrule{12-15}\cmidrule{17-20}          & \multicolumn{2}{c}{$K$=50} & \multicolumn{2}{c}{$K$=100} &       & \multicolumn{2}{c}{$K$=50} & \multicolumn{2}{c}{$K$=100} &       & \multicolumn{2}{c}{$K$=50} & \multicolumn{2}{c}{$K$=100} &       & \multicolumn{2}{c}{$K$=50} & \multicolumn{2}{c}{$K$=100} \\
            & \multicolumn{1}{c}{Purity} & \multicolumn{1}{c}{NMI} & \multicolumn{1}{c}{Purity} & \multicolumn{1}{c}{NMI} &       & \multicolumn{1}{c}{Purity} & \multicolumn{1}{c}{NMI} & \multicolumn{1}{c}{Purity} & \multicolumn{1}{c}{NMI} &       & \multicolumn{1}{c}{Purity} & \multicolumn{1}{c}{NMI} & \multicolumn{1}{c}{Purity} & \multicolumn{1}{c}{NMI} &       & \multicolumn{1}{c}{Purity} & \multicolumn{1}{c}{NMI} & \multicolumn{1}{c}{Purity} & \multicolumn{1}{c}{NMI} \\
      \midrule
      LDA   & $^\ddag$0.367 & $^\ddag$0.364 & $^\ddag$0.364 & $^\ddag$0.346 &       & $^\ddag$0.614 & $^\ddag$0.041 & $^\ddag$0.600 & $^\ddag$0.037 &       & $^\ddag$0.288 & $^\ddag$0.144 & $^\ddag$0.297 & $^\ddag$0.148 &       & $^\ddag$0.640 & $^\ddag$0.193 & $^\ddag$0.654 & $^\ddag$0.194 \\
      WLDA  & $^\ddag$0.233 & $^\ddag$0.157 & $^\ddag$0.292 & $^\ddag$0.207 &       & $^\ddag$0.589 & $^\ddag$0.011 & $^\ddag$0.602 & $^\ddag$0.013 &       & $^\ddag$0.255 & $^\ddag$0.084 & $^\ddag$0.303 & $^\ddag$0.127 &       & $^\ddag$0.580 & $^\ddag$0.151 & $^\ddag$0.653 & $^\ddag$0.188 \\
      DVAE  & $^\ddag$0.087 & $^\ddag$0.018 & $^\ddag$0.104 & $^\ddag$0.035 &       & $^\ddag$0.517 & $^\ddag$0.000 & $^\ddag$0.525 & $^\ddag$0.001 &       & $^\ddag$0.171 & $^\ddag$0.030 & $^\ddag$0.202 & $^\ddag$0.057 &       & $^\ddag$0.713 & $^\ddag$0.219 & $^\ddag$0.407 & $^\ddag$0.030 \\
      ETM   & $^\ddag$0.347 & $^\ddag$0.319 & $^\ddag$0.394 & $^\ddag$0.339 &       & $^\ddag$0.660 & $^\ddag$0.038 & $^\ddag$0.648 & $^\ddag$0.037 &       & $^\ddag$0.405 & $^\ddag$0.192 & $^\ddag$0.428 & $^\ddag$0.208 &       & $^\ddag$0.679 & $^\ddag$0.224 & $^\ddag$0.674 & $^\ddag$0.204 \\
      HyperMiner & $^\ddag$0.433 & $^\ddag$0.405 & $^\ddag$0.454 & $^\ddag$0.386 &       & $^\ddag$0.655 & $^\ddag$0.046 & $^\ddag$0.641 & $^\ddag$0.032 &  & $^\ddag$0.456 & $^\ddag$0.237 & $^\ddag$0.448 & $^\ddag$0.222 &       & $^\ddag$0.730 & $^\ddag$0.276 & $^\ddag$0.679 & $^\ddag$0.200 \\

      NSTM  & $^\ddag$0.354 & $^\ddag$0.356 & $^\ddag$0.383 & $^\ddag$0.363 &       & $^\ddag$0.658 & $^\ddag$0.040 & $^\ddag$0.659 & $^\ddag$0.039 &       & $^\ddag$0.395 & $^\ddag$0.241 & $^\ddag$0.405 & $^\ddag$0.242 &       & $^\ddag$0.719 & $^\ddag$0.324 & $^\ddag$0.764 & $^\ddag$0.359 \\
      WeTe  & $^\ddag$0.268 & $^\ddag$0.304 & $^\ddag$0.338 & $^\ddag$0.348 &       & $^\ddag$0.587 & $^\ddag$0.031 & $^\ddag$0.589 & $^\ddag$0.025 &       & $^\ddag$0.389 & $^\ddag$0.252 & $^\ddag$0.444 & $^\ddag$0.269 &       & $^\ddag$0.641 & $^\ddag$0.268 & $^\ddag$0.699 & $^\ddag$0.271 \\
      \midrule
      \textbf{\modelname{}} & \textbf{0.560} & \textbf{0.524} & \textbf{0.555} & \textbf{0.494} &       & \textbf{0.694} & \textbf{0.058} & \textbf{0.694} & \textbf{0.049} &       & \textbf{0.550} & \textbf{0.295} & \textbf{0.563} & \textbf{0.311} &       & \textbf{0.802} & \textbf{0.367} & \textbf{0.812} & \textbf{0.428} \\
      \bottomrule
      \end{tabular}%
      }
    \caption{
        Document clustering of Purity and NMI under 50 and 100 topics ($K\!\!=\!\!50$ and $K\!\!=\!\!100$). The best scores are in \textbf{bold}.
        $\ddag$ means the gain of \modelname{} is statistically significant at 0.05 level.
    }
    \label{tab_clustering}%
  \end{table*}%

\section{Experiment}

    \subsection{Experiment Setup}
        \paragraph{Datasets}
            We adopt the following benchmark document datasets for experiments:
            \begin{inparaenum}[(i)]
                \item 20 News Groups \citep[\textbf{20NG},][]{Lang95} 
                is one of the most popular datasets for evaluating topic models, including news articles with 20 labels;
                \item \textbf{IMDB} \cite{maas2011learning} 
                is the movie reviews containing two labels (positive and negative);
                \item \textbf{Yahoo Answer} \cite{Zhang2015} is the question titles, contents, and the best answers from the Yahoo website with 10 labels, such as Society, Culture, and Family \& Relationships;
                \item \textbf{AG News} \cite{Zhang2015} 
                contains news titles and descriptions, divided into 4 categories like Sports and Business.
            \end{inparaenum}
            Note that Yahoo Answer and AG News belong to short texts.
            See \Cref{sec_appendix_dataset} for pre-processing details.

        \paragraph{Evaluation Metrics}
            Following previous mainstream work,
            we evaluate topic models concerning \textbf{topic quality and doc-topic distribution quality}.
            Topic quality includes:
            \begin{inparaenum}[(i)]
                \item
                    \textbf{Topic Coherence} measures the coherence between the top words of discovered topics \cite{Newman2010,Wang2011}.
                    We employ the widely-used metric, Coherence Value ($C_V$) which has been empirically shown to outperform the traditional metrics, NPMI, UCI, and UMass \cite{roder2015exploring}.
                    We also exemplify this in \Cref{sec_appendix_comparison_coherence_metrics}.
                    We use the public Wikipedia article collection~\footnote{\url{https://github.com/dice-group/Palmetto}} as the external reference corpus.
                    This removes the bias of using relatively small datasets (\emph{e.g.}, training sets) as the reference corpus,
                    so we can reach fair comparisons and good reproducibility.
                \item
                    \textbf{Topic Diversity} measures the differences between discovered topics to verify if topic collapsing happens.
                    We use the Topic Diversity metric \citep[TD,][]{dieng2020topic} to evaluate this performance, which computes the proportion of unique words in the discovered topics.
            \end{inparaenum}
            We select the top 15 words of discovered topics for the above topic quality evaluation.
            We furthermore conduct \textbf{document clustering} experiments to evaluate doc-topic distribution quality with Purity and NMI following \citet{zhao2020neural,wang2022representing}.

        \paragraph{Baseline Models}
            We consider the following state-of-the-art models for comparison:
            \begin{inparaenum}[(i)]
                \item \textbf{LDA} \cite{blei2003latent}, one of the most widely-used probabilistic topic models;
                \item
                    \textbf{KM} \cite{Sia2020}, directly clustering word embeddings to produce topics.
                    Note that we cannot use it for document clustering since it cannot infer the doc-topic distributions.
                \item
                    \textbf{DVAE} \cite{burkhardt2019decoupling}, Dirichlet VAE that approximates Dirichlet priors
                    with rejection sampling;
                \item
                    \textbf{WLDA} \cite{Nan2019}, a WAE-based topic model;
                \item
                    \textbf{ETM} \cite{dieng2020topic}, a neural topic model which models the topic-word distribution matrix with word and topic embeddings;
                \item
                    \textbf{HyperMiner} \cite{xu2022hyperminer},
                    using embeddings in the hyperbolic space to model topics.
                \item
                    \textbf{NSTM} \cite{zhao2020neural}, using optimal transport distance between doc-topic distributions and documents to measure reconstruction error.
                \item
                    \textbf{WeTe} \cite{wang2022representing}, following NSTM and using conditional transport distance as reconstruction error. 
            \end{inparaenum}

    \subsection{Topic and Doc-Topic Distribution Quality} \label{sec_main_results}
            \Cref{tab_topic_quality} reports
            the topic quality results concerning $C_V$ and TD, and \Cref{tab_clustering} summarizes doc-topic distribution quality results concerning Purity and NMI of document clustering.
            We have the following observations:
            \begin{inparaenum}[(\bgroup\bfseries i\egroup)]
                \item
                    \textbf{
                        \modelname{} effectively addresses the topic collapsing issue and outperforms baselines in topic quality.
                    }
                    In \Cref{tab_topic_quality},
                    the much lower TD scores imply baselines generate repetitive topics and thus suffer from topic collapsing.
                    As aforementioned, these repetitive topics are less useful for downstream tasks and damage the interpretability of topic models.
                    In contrast, we see our \modelname{} achieves significantly higher TD scores across all datasets
                    and mostly the best $C_V$ scores.
                    We emphasize although the $C_V$ of \modelname{} is slightly higher than NSTM (0.389 v.s. 0.387) on Yahoo Answer,
                    \modelname{} completely outperforms on TD (0.903 v.s. 0.659).
                    These results demonstrate that \modelname{} produces more coherent and diverse topics than state-of-the-art baselines.
                    These improvements are because
                    our \modelname{} makes topic embeddings away from each other and cover different semantics of word embeddings in the space instead of collapsing together as some baselines.
                \item
                    \textbf{\modelname{} surpasses baselines in inferring high-quality doc-topic distributions.}
                    \Cref{tab_clustering} shows our \modelname{} consistently outperforms the baseline models by a large margin in terms of Purity and NMI.
                    For example, \modelname{} reaches 0.560 and 0.524 for Purity and NMI on 20NG, while the runner-up only has 0.367 and 0.364.
                    These manifest that \modelname{} not only achieves higher-quality topics but also better doc-topic distributions as document representations.
            \end{inparaenum}
            \textbf{See \Cref{sec_appendix_perplexity,sec_appendix_range,sec_appendix_comparison_coherence_metrics,sec_appendix_visual}
            for more experiments like robustness to the number of topics and visualization results.}

\begin{figure*}[!ht]
    \centering
    \begin{minipage}[b]{0.5\textwidth}
    \centering
    \setlength{\tabcolsep}{1.3mm}
    \resizebox{\linewidth}{!}{
        \begin{tabular}[b]{lrrrrrrrrr}
        \toprule
        \multicolumn{1}{c}{\multirow{2}[4]{*}{Model}} & \multicolumn{4}{c}{20NG}      &       & \multicolumn{4}{c}{Yahoo Answer} \\
        \cmidrule{2-5}\cmidrule{7-10}      & \multicolumn{1}{c}{Purity} & \multicolumn{1}{c}{NMI} & \multicolumn{1}{c}{$C_V$} & \multicolumn{1}{c}{TD} &       & \multicolumn{1}{c}{Purity} & \multicolumn{1}{c}{NMI} & \multicolumn{1}{c}{$C_V$} & \multicolumn{1}{c}{TD} \\
        \midrule
        DKM   & $^\ddag$0.510 & $^\ddag$0.471 & 0.448 & $^\ddag$0.577 &       & $^\ddag$0.507 & $^\ddag$0.282 & 0.403 & $^\ddag$0.631 \\
        DKM+Entropy & $^\ddag$0.222 & $^\ddag$0.148 & 0.469 & $^\ddag$0.503 &       & $^\ddag$0.252 & $^\ddag$0.092 & 0.433 & $^\ddag$0.592 \\
        w/o ECR & $^\ddag$0.504 & $^\ddag$0.446 & 0.461 & $^\ddag$0.548 &       & $^\ddag$0.498 & $^\ddag$0.262 & 0.435 & $^\ddag$0.608 \\
        \midrule
        \textbf{ECRTM} & 0.560 & 0.524 & 0.431 & 0.964 &       & 0.550 & 0.295 & 0.405 & 0.985 \\
        \bottomrule
        \end{tabular}%
    }
    \captionof{table}{
        Ablation study.
        The terrible TD scores of DKM, DKM+Entropy and w/o \regname{}
        indicate topic collapsing still exists, making their high $C_V$ less meaningful.
        In contrast, \modelname{} achieves much higher TD with the best Purity and NMI.
        The $C_V$ scores of \modelname{} also outperforms state-of-the-art baselines (See \Cref{tab_topic_quality}).
        $\ddag$~means the gain of \modelname{} is statistically significant at 0.05 level.
    }
    \label{tab_ablation}
    \end{minipage}%
    \hfill
    \begin{minipage}[b]{0.45\textwidth}
        \centering
        \includegraphics[width=\linewidth]{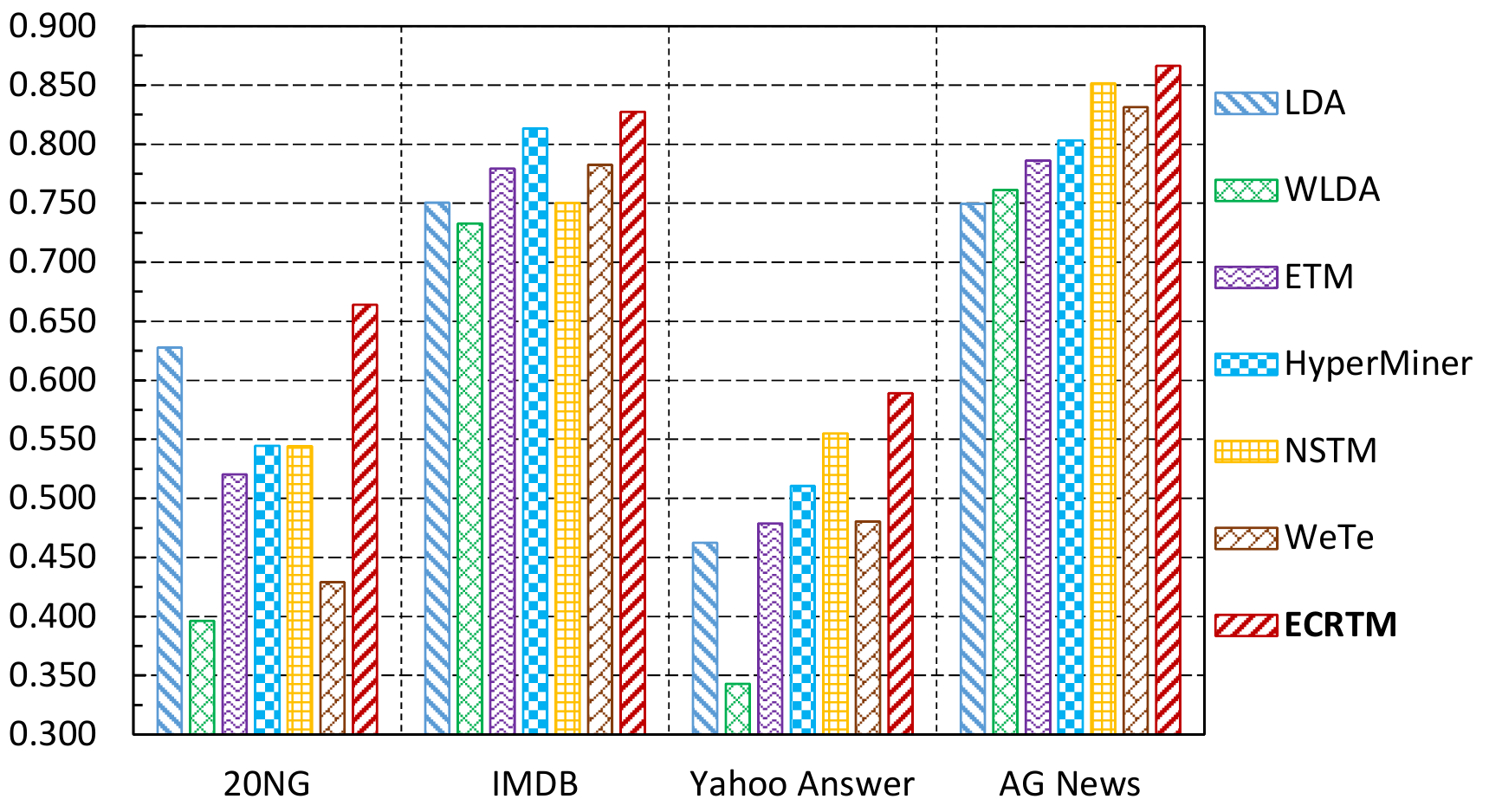}
        \caption{
            Text classification results of F1 scores.
            The improvements of \modelname{} are all statistically significant at 0.05 level.
        }
    \label{fig_classification}
    \end{minipage}%
\end{figure*}

\begin{table*}[!ht]
        \centering
        \setlength{\tabcolsep}{7mm}
        \resizebox{0.75\linewidth}{!}{
            \begin{tabular}{ll}
                \toprule
                \multirow{3}[2]{*}{ETM}
                    & \uline{like} better \uline{good} especially end \uline{look} \uline{much} done \uline{way} \uline{just} \\
                    & \uline{like} \uline{just} one \uline{way} made \uline{much} times really even feel \\
                    & one \uline{like} around sort looking kind \uline{good} main \uline{look} \uline{just} \\
                \midrule
                \multirow{3}[2]{*}{HyperMiner}
                    & \uline{one} \uline{even} \uline{end} \uline{way} \uline{little} \uline{part} \uline{character} \uline{make} never \uline{plot} \\
                    & \uline{even} \uline{end} \uline{little} \uline{way} \uline{one} \uline{plot} \uline{character} enough \uline{part} \uline{make} \\
                    & \uline{even} seems fact enough \uline{plot} \uline{end} least \uline{character} audience \uline{make} \\
                \midrule
                \multirow{3}[2]{*}{NSTM}
                    & \uline{just} show \uline{even} \uline{come} time one \uline{good} \uline{really} going know \\
                    & \uline{just} \uline{even} \uline{really} \uline{something} \uline{come} going \uline{like} actually things get \\
                    & \uline{just} one \uline{even} \uline{something} \uline{come} way \uline{really} \uline{like} always \uline{good} \\
                \midrule
                \multirow{3}[2]{*}{WeTe}
                    & just like really bad good get one think see even \\
                    & man back gets goes two get takes house around away \\
                    & jokes jackson lisa predictable recycled wasted murphy writers williams rock \\
                \midrule
                \multirow{3}[2]{*}{DKM}
                    & \uline{christmas} disney musical songs bill timeless prince art rock \uline{holiday} \\
                    & \uline{christmas} \uline{santa} \uline{childrens} \uline{holiday} betty age ann adult children toy \\
                    & fantasy \uline{christmas} magic effects magical \uline{santa} special \uline{holiday} \uline{childrens} child \\
                \midrule
                \multirow{3}[2]{*}{DKM+Entropy}
                    & \uline{funny} \uline{day} \uline{physical} \uline{semi} \uline{ever} \uline{way} \uline{old} \uline{due} \uline{seen} zone \\
                    & \uline{funny} \uline{ever} \uline{day} \uline{old} \uline{seen} \uline{way} \uline{physical} \uline{due} \uline{semi} relationship \\
                    & \uline{funny} \uline{ever} \uline{day} \uline{semi} \uline{seen} \uline{physical} \uline{way} \uline{old} \uline{due} psychological \\
                \midrule
                \multirow{3}[2]{*}{\textbf{\modelname{}}}
                    & jackie martial chan kung arts kong hong stunts bruce fight \\
                    & nominated nancy academy award awards oscar oscars jake nomination dracula  \\
                    & vampires vampire freddy zombies zombie nightmare serial halloween killer slasher \\
                \bottomrule
            \end{tabular}%
        }
        \caption{
            Case study: each row is the top 10 related words of a discovered topic.
            Repetitive words are \uline{underlined}.
        }
        \label{tab_topic_example}
\end{table*}

    \subsection{Ablation Study} \label{sec_ablation}
        We conduct ablation studies and show the necessity of our proposed \regfullname{} (\regname{}).
        Specifically,
        we remove the \regname{} from our \modelname{}, denoted as w/o \regname{}.
        We also compare with the state-of-the-art deep clustering method, DKM \cite{fard2020deep} and DKM with minimizing entropy (DKM+Entropy, see \Cref{sec_how_to}).
        Note that we are the first to use DKM in topic modeling.
        \Cref{tab_ablation} shows DKM, DKM+Entropy, and w/o \regname{} all suffer from topic collapsing as indicated by their much lower TD scores.
        Although they have high $C_V$, their terrible TD scores
        mean most topics are repetitive and less useful for downstream tasks, making their high $C_V$ scores less meaningful (see examples in \Cref{sec_appendix_topic_examples} for illustrations).
        Conversely, our \modelname{} improves TD scores by a large margin and achieves the best document clustering performance with much higher Purity and NMI.
        This is because our \regname{}, as an effective regularization, can avoid the collapsing of topic embeddings while DKM, DKM+Entropy, and w/o \regname{} cannot.
        These results demonstrate our \regname{} is necessary to address the topic collapsing issue and achieve effective topic modeling performance.

\begin{table*}[!ht]
    \centering
    \setlength{\tabcolsep}{1.5mm}
    \renewcommand{\arraystretch}{1.2}
    \resizebox{\linewidth}{!}{
        \begin{tabular}{lrrrrrrrrrrrrrrrrrrrrrrrr}
        \toprule
        \multirow{2}[4]{*}{Model} & \multicolumn{4}{c}{max-df=0.5} &       & \multicolumn{4}{c}{max-df=0.4} &       & \multicolumn{4}{c}{max-df=0.3} &       & \multicolumn{4}{c}{max-df=0.2} &       & \multicolumn{4}{c}{max-df=0.1} \\
        \cmidrule{2-5}\cmidrule{7-10}\cmidrule{12-15}\cmidrule{17-20}\cmidrule{22-25}      & \multicolumn{1}{c}{Purity} & \multicolumn{1}{c}{NMI} & \multicolumn{1}{c}{$C_V$} & \multicolumn{1}{c}{TD} &       & \multicolumn{1}{c}{Purity} & \multicolumn{1}{c}{NMI} & \multicolumn{1}{c}{$C_V$} & \multicolumn{1}{c}{TD} &       & \multicolumn{1}{c}{Purity} & \multicolumn{1}{c}{NMI} & \multicolumn{1}{c}{$C_V$} & \multicolumn{1}{c}{TD} &       & \multicolumn{1}{c}{Purity} & \multicolumn{1}{c}{NMI} & \multicolumn{1}{c}{$C_V$} & \multicolumn{1}{c}{TD} &       & \multicolumn{1}{c}{Purity} & \multicolumn{1}{c}{NMI} & \multicolumn{1}{c}{$C_V$} & \multicolumn{1}{c}{TD} \\
        \midrule
        LDA   & 0.402 & 0.375 & 0.378 & 0.661 &       & 0.395 & 0.372 & 0.367 & 0.685 &       & 0.431 & 0.394 & 0.379 & 0.724 &       & 0.403 & 0.407 & 0.379 & 0.733 &       & 0.460 & 0.428 & 0.389 & 0.820 \\
        KM    & /     & /     & 0.264 & 0.193 &       & /     & /     & 0.277 & 0.260 &       & /     & /     & 0.270 & 0.197 &       & /     & /     & 0.267 & 0.240 &       & /     & /     & 0.260 & 0.203 \\
        WLDA  & 0.221 & 0.154 & 0.353 & 0.409 &       & 0.223 & 0.144 & 0.343 & 0.413 &       & 0.222 & 0.150 & 0.351 & 0.388 &       & 0.232 & 0.153 & 0.365 & 0.420 &       & 0.243 & 0.167 & 0.376 & 0.473 \\
        HyperMiner & 0.465 & 0.429 & 0.360 & 0.600 &       & 0.430 & 0.398 & 0.356 & 0.583 &       & 0.445 & 0.425 & 0.363 & 0.599 &       & 0.457 & 0.434 & 0.369 & 0.577 &       & 0.495 & 0.444 & 0.375 & 0.708 \\
        ETM   & 0.365 & 0.330 & 0.384 & 0.736 &       & 0.367 & 0.301 & 0.367 & 0.735 &       & 0.354 & 0.307 & 0.372 & 0.735 &       & 0.367 & 0.307 & 0.379 & 0.760 &       & 0.370 & 0.319 & 0.376 & 0.844 \\
        NSTM  & 0.351 & 0.351 & 0.398 & 0.435 &       & 0.383 & 0.382 & 0.388 & 0.423 &       & 0.447 & 0.409 & 0.394 & 0.432 &       & 0.396 & 0.388 & 0.399 & 0.511 &       & 0.393 & 0.365 & 0.405 & 0.620 \\
        WeTe  & 0.381 & 0.422 & 0.380 & 0.944 &       & 0.299 & 0.349 & 0.385 & 0.961 &       & 0.311 & 0.387 & 0.371 & 0.923 &       & 0.326 & 0.395 & 0.384 & 0.957 &       & 0.376 & 0.450 & 0.394 & 0.957 \\
        \midrule
        \textbf{ECRTM} & \textbf{0.556} & \textbf{0.525} & \textbf{0.431} & \textbf{0.992} &       & \textbf{0.516} & \textbf{0.503} & \textbf{0.417} & \textbf{0.992} &       & \textbf{0.599} & \textbf{0.556} & \textbf{0.436} & \textbf{0.952} &       & \textbf{0.579} & \textbf{0.524} & \textbf{0.432} & \textbf{0.987} &       & \textbf{0.546} & \textbf{0.514} & \textbf{0.416} & \textbf{0.976} \\
        \bottomrule
        \end{tabular}%
    }
    \caption{
        Influence of high-frequency words.
        Here max-df denotes the maximum document frequency of dataset pre-processing.
        A smaller max-df removes more high-frequency words. 
        The best scores are in \textbf{bold}.
    }
  \label{tab_maxdf}%
\end{table*}%

    \subsection{Text Classification}
        To evaluate extrinsically, we further conduct text classification experiments as downstream tasks.
        Specifically, we use the doc-topic distributions learned by topic models as document features and train SVMs to predict the class of each document.
        As reported in \Cref{fig_classification}, \modelname{} significantly outperforms baseline models on all datasets.
        These results demonstrate that
        our \modelname{} can be better utilized in the downstream classification tasks.

    \subsection{Case Study: Examples of Discovered Topics} \label{sec_appendix_topic_examples}
        For case study, \Cref{tab_topic_example} shows examples of discovered topics by different models from IMDB.
        We observe that ETM and NSTM both have highly uninformative and similar topics including common words like ``just'', ``like'', or ``something''.
        HyperMiner generates repetitive topics with the words ``one'', ``even'', and ``end''
        WeTe produces some less informative topics like ``just like really bad good...''.
        DKM and DKM+Entropy also have repetitive topics with the words ``christmas'', ``holiday'', and ``funny''.
        Accordingly, we observe that the topic collapsing issue commonly exists in these methods.
        These collapsed topics are uninformative and redundant, which are less useful for downstream applications and damage the interpretability of topic models.
        In contrast, the topics discovered by \modelname{} are more distinct instead of repeating each other.
        Besides, they are more coherent, such as the first topic with relevant words like ``jackie'', ``chan'', and ``stunts''.
        \Cref{sec_appendix_full_list} shows the full topics lists of models.

    \subsection{Influence of Dataset Pre-processing} \label{sec_high-frequency}
        As aforementioned in \Cref{sec_why},
        we argue that topic collapsing results from the reconstruction error minimization on high-frequency words.
        This inspires us to ask: what if we carefully remove high-frequency words using reliable dataset pre-processing?
        Driven by this,
        we alter the maximum document frequency (max-df) to remove the high-frequency words when pre-processing the 20NG dataset.
        A smaller max-df removes more high-frequency words.
        From \Cref{tab_maxdf}
        we see that
        most baselines, such as LDA, ETM, and NSTM, reach higher TD scores under small max-df, indicating that topic collapsing is alleviated to some extent.
        But our model consistently outperforms all baselines on the topic quality and document clustering.
        These results empirically confirm our argument that the topic collapsing issue arises from reconstructing high-frequency words.
        Besides, these results verify one of our advantages:
        \textbf{our model requires no reliable pre-processing to achieve state-of-the-art performance}.
        This advantage is vital since the definition of reliable pre-processing is inconclusive:
        A large max-df may not remove any high-frequency words while a small one may remove most of the important words.
        More critically, brutally searching for reliable pre-processing is time-consuming and laborious.
        This advantage becomes more significant when meeting many large-scale datasets from various domains.

\section{Conclusion}
    In this paper, we propose the novel \modelfullname{} (\modelname{}) to address the topic collapsing issue.
    \modelname{}
    learns topics under the new \regfullname{} that forces each topic embedding to be the center of a separately aggregated word embedding cluster.
    Extensive experiments demonstrate that \modelname{} achieves effective neural topic modeling, successfully alleviates topic collapsing, and consistently achieves state-of-the-art performance in terms of producing high-quality topics and topic distributions of documents.

\section*{Acknowledgements}
    We thank all anonymous reviewers for their helpful comments.
    This research/project is supported by the National Research Foundation, Singapore under its AI Singapore Programme, AISG Award No: AISG2-TC-2022-005 and AISG Award No: AISG-100E-2019-046.

\bibliographystyle{icml2023}
\bibliography{library}

\newpage
\appendix

\clearpage

\begin{table*}[!ht]
    \centering
    \setlength{\tabcolsep}{1.5mm}
    \renewcommand{\arraystretch}{1.2}
    \resizebox{\linewidth}{!}{
    \begin{tabular}{lrrrrrrrrrrrrrrrrrrrrrrr}
    \toprule
    \multirow{2}[4]{*}{Model} & \multicolumn{2}{c}{$K$=10} &       & \multicolumn{2}{c}{$K$=20} &       & \multicolumn{2}{c}{$K$=30} &       & \multicolumn{2}{c}{$K$=40} &       & \multicolumn{2}{c}{$K$=60} &       & \multicolumn{2}{c}{$K$=70} &       & \multicolumn{2}{c}{$K$=80} &       & \multicolumn{2}{c}{$K$=90} \\
    & \multicolumn{1}{c}{$C_V$} & \multicolumn{1}{c}{TD} &       & \multicolumn{1}{c}{$C_V$} & \multicolumn{1}{c}{TD} &       & \multicolumn{1}{c}{$C_V$} & \multicolumn{1}{c}{TD} &       & \multicolumn{1}{c}{$C_V$} & \multicolumn{1}{c}{TD} &       & \multicolumn{1}{c}{$C_V$} & \multicolumn{1}{c}{TD} &       & \multicolumn{1}{c}{$C_V$} & \multicolumn{1}{c}{TD} &       & \multicolumn{1}{c}{$C_V$} & \multicolumn{1}{c}{TD} &       & \multicolumn{1}{c}{$C_V$} & \multicolumn{1}{c}{TD} \\
    \midrule
    LDA   & 0.376 & 0.627 &       & 0.384 & 0.620 &       & 0.377 & 0.678 &       & 0.390 & 0.697 &       & 0.373 & 0.674 &       & 0.385 & 0.590 &       & 0.381 & 0.645 &       & 0.380 & 0.656 \\
    KM    & 0.208 & 0.207 &       & 0.230 & 0.180 &       & 0.247 & 0.151 &       & 0.255 & 0.217 &       & 0.269 & 0.219 &       & 0.282 & 0.256 &       & 0.300 & 0.321 &       & 0.289 & 0.300 \\
    WLDA  & 0.354 & 0.533 &       & 0.354 & 0.443 &       & 0.360 & 0.389 &       & 0.357 & 0.430 &       & 0.361 & 0.343 &       & 0.373 & 0.335 &       & 0.371 & 0.345 &       & 0.375 & 0.287 \\
    ETM   & 0.380 & 0.820 &       & 0.372 & 0.763 &       & 0.373 & 0.744 &       & 0.383 & 0.698 &       & 0.367 & 0.672 &       & 0.377 & 0.667 &       & 0.377 & 0.663 &       & 0.372 & 0.629 \\
    HyperMiner & 0.361 & 0.800 &       & 0.377 & 0.750 &       & 0.373 & 0.671 &       & 0.378 & 0.665 &       & 0.379 & 0.590 &       & 0.380 & 0.533 &       & 0.373 & 0.516 &       & 0.369 & 0.454 \\
    NSTM  & 0.397 & 0.600 &       & 0.381 & 0.487 &       & 0.385 & 0.391 &       & 0.389 & 0.418 &       & 0.392 & 0.444 &       & 0.396 & 0.520 &       & 0.400 & 0.468 &       & 0.386 & 0.450 \\
    WeTe  & 0.422 & \textbf{1.000} &       & 0.380 & 0.980 &       & 0.387 & 0.980 &       & 0.388 & 0.978 &       & 0.378 & 0.948 &       & 0.368 & 0.879 &       & 0.355 & 0.800 &       & 0.349 & 0.754 \\
    \midrule
    \textbf{\modelname{}} & \textbf{0.487} & \textbf{1.000} &       & \textbf{0.454} & \textbf{1.000} &       & \textbf{0.437} & \textbf{1.000} &       & \textbf{0.435} & \textbf{0.993} &       & \textbf{0.413} & \textbf{0.993} &       & \textbf{0.405} & \textbf{0.910} &       & \textbf{0.410} & \textbf{0.957} &       & \textbf{0.402} & \textbf{0.906} \\
    \bottomrule
    \end{tabular}%
    }
    \caption{
        Topic quality of coherence ($C_V$) and diversity (TD) under topic number $K\!\!=\!\!10,20,30,40,60,70,80,90$. 
        The best scores are in \textbf{bold}.
    }
    \label{tab_range_topic_quality}
\end{table*}%

\begin{table*}[!ht]
    \centering
    \setlength{\tabcolsep}{1.5mm}
    \renewcommand{\arraystretch}{1.2}
    \resizebox{\linewidth}{!}{
    \begin{tabular}{lrrrrrrrrrrrrrrrrrrrrrrr}
        \toprule
        \multirow{2}[4]{*}{Model} & \multicolumn{2}{c}{$K$=10} &       & \multicolumn{2}{c}{$K$=20} &       & \multicolumn{2}{c}{$K$=30} &       & \multicolumn{2}{c}{$K$=40} &       & \multicolumn{2}{c}{$K$=60} &       & \multicolumn{2}{c}{$K$=70} &       & \multicolumn{2}{c}{$K$=80} &       & \multicolumn{2}{c}{$K$=90} \\
        \cmidrule{2-3}\cmidrule{5-6}\cmidrule{8-9}\cmidrule{11-12}\cmidrule{14-15}\cmidrule{17-18}\cmidrule{20-21}\cmidrule{23-24}      & \multicolumn{1}{c}{Purity} & \multicolumn{1}{c}{NMI} &       & \multicolumn{1}{c}{Purity} & \multicolumn{1}{c}{NMI} &       & \multicolumn{1}{c}{Purity} & \multicolumn{1}{c}{NMI} &       & \multicolumn{1}{c}{Purity} & \multicolumn{1}{c}{NMI} &       & \multicolumn{1}{c}{Purity} & \multicolumn{1}{c}{NMI} &       & \multicolumn{1}{c}{Purity} & \multicolumn{1}{c}{NMI} &       & \multicolumn{1}{c}{Purity} & \multicolumn{1}{c}{NMI} &       & \multicolumn{1}{c}{Purity} & \multicolumn{1}{c}{NMI} \\
        \midrule
        LDA   & 0.295 & 0.408 &       & 0.340 & 0.396 &       & 0.347 & 0.375 &       & 0.368 & 0.356 &       & 0.354 & 0.352 &       & 0.399 & 0.368 &       & 0.378 & 0.352 &       & 0.389 & 0.359 \\
        WLDA  & 0.174 & 0.119 &       & 0.194 & 0.124 &       & 0.223 & 0.152 &       & 0.238 & 0.161 &       & 0.260 & 0.176 &       & 0.272 & 0.186 &       & 0.252 & 0.180 &       & 0.281 & 0.199 \\
        ETM   & 0.183 & 0.274 &       & 0.275 & 0.307 &       & 0.307 & 0.288 &       & 0.331 & 0.281 &       & 0.351 & 0.291 &       & 0.340 & 0.302 &       & 0.379 & 0.330 &       & 0.407 & 0.349 \\
        HyperMiner & 0.240 & 0.299 &       & 0.338 & 0.390 &       & 0.416 & 0.421 &       & 0.407 & 0.389 &       & 0.478 & 0.422 &       & 0.461 & 0.408 &       & 0.468 & 0.396 &       & 0.446 & 0.384 \\
        NSTM  & 0.228 & 0.284 &       & 0.295 & 0.327 &       & 0.355 & 0.373 &       & 0.349 & 0.349 &       & 0.362 & 0.353 &       & 0.357 & 0.344 &       & 0.351 & 0.365 &       & 0.376 & 0.354 \\
        WeTe  & 0.055 & 0.004 &       & 0.119 & 0.150 &       & 0.197 & 0.244 &       & 0.252 & 0.317 &       & 0.281 & 0.332 &       & 0.384 & 0.421 &       & 0.313 & 0.331 &       & 0.302 & 0.311 \\
        \midrule
        \textbf{\modelname{}} & \textbf{0.390} & \textbf{0.485} &       & \textbf{0.373} & \textbf{0.420} &       & \textbf{0.463} & \textbf{0.435} &       & \textbf{0.462} & \textbf{0.426} &       & \textbf{0.554} & \textbf{0.522} &       & \textbf{0.559} & \textbf{0.498} &       & \textbf{0.581} & \textbf{0.506} &       & \textbf{0.564} & \textbf{0.497} \\
        \bottomrule
    \end{tabular}%
    }
    \caption{
        Document clustering of Purity and NMI under topic number $K\!\!=\!\!10,20,30,40,60,70,80,90$. 
        The best scores are in \textbf{bold}.
    }
    \label{tab_range_clustering}
\end{table*}%

\appendix

\section{Dataset} \label{sec_appendix_dataset}
    We follow the dataset pre-processing steps of \cite{Card2018a}:
    \begin{inparaenum}[(1)]
        \item tokenize documents and convert to lowercase;
        \item remove punctuation;
        \item remove tokens that include numbers;
        \item remove tokens less than 3 characters;
        \item remove stop words.
    \end{inparaenum}
    The statistics of pre-processed datasets are reported in \Cref{tab_dataset}.

\begin{table}[!t]
    \centering
    \resizebox{0.95\linewidth}{!}{
    \begin{tabular}{lrrrr}
    \toprule
    Dataset & \#docs & \makecell[r]{Vocabulary \\ Size} & \makecell[r]{Average \\ Length} & \#labels \\
    \midrule
    20NG  & 18,846  & 5,000  & 110.5 & 20 \\
    IMDB  & 50,000  & 5,000  & 95.0  & 2 \\
    Yahoo Answer & 12,500  & 5,000  & 35.4  & 10 \\
    AG News & 12,500  & 5,000  & 20.1  & 4 \\
    \bottomrule
    \end{tabular}%
    }
    \caption{Statistics of datasets after pre-processing.}
    \label{tab_dataset}
\end{table}

\begin{table}[!t]
    \centering
    \setlength{\tabcolsep}{2mm}
    \renewcommand{\arraystretch}{1.1}
    \resizebox{0.9\linewidth}{!}{
        \begin{tabular}{lcccc}
        \toprule
        Model & 20NG & IMDB & Yahoo Answer & AG News \\
        \midrule
        LDA   & 2044.6 & 2482.8 & 4637.1 & 9951.1 \\
        DVAE  & 2045.0 & 1996.3 & 3258.3 & 1851.3 \\
        ETM   & 2113.4 & 1911.8 & 4653.4 & 1518.7 \\
        \midrule
        \textbf{\modelname{}} & \textbf{1896.9} & \textbf{1830.8} & \textbf{3244.0} & \textbf{1436.1} \\
        \bottomrule
        \end{tabular}%
    }
    \caption{Perplexity results. The best is in \textbf{bold} (lower is better).}
    \label{tab_perplexity}%
\end{table}%

\begin{figure*}[!ht]
    \centering
    \includegraphics[width=0.6\linewidth]{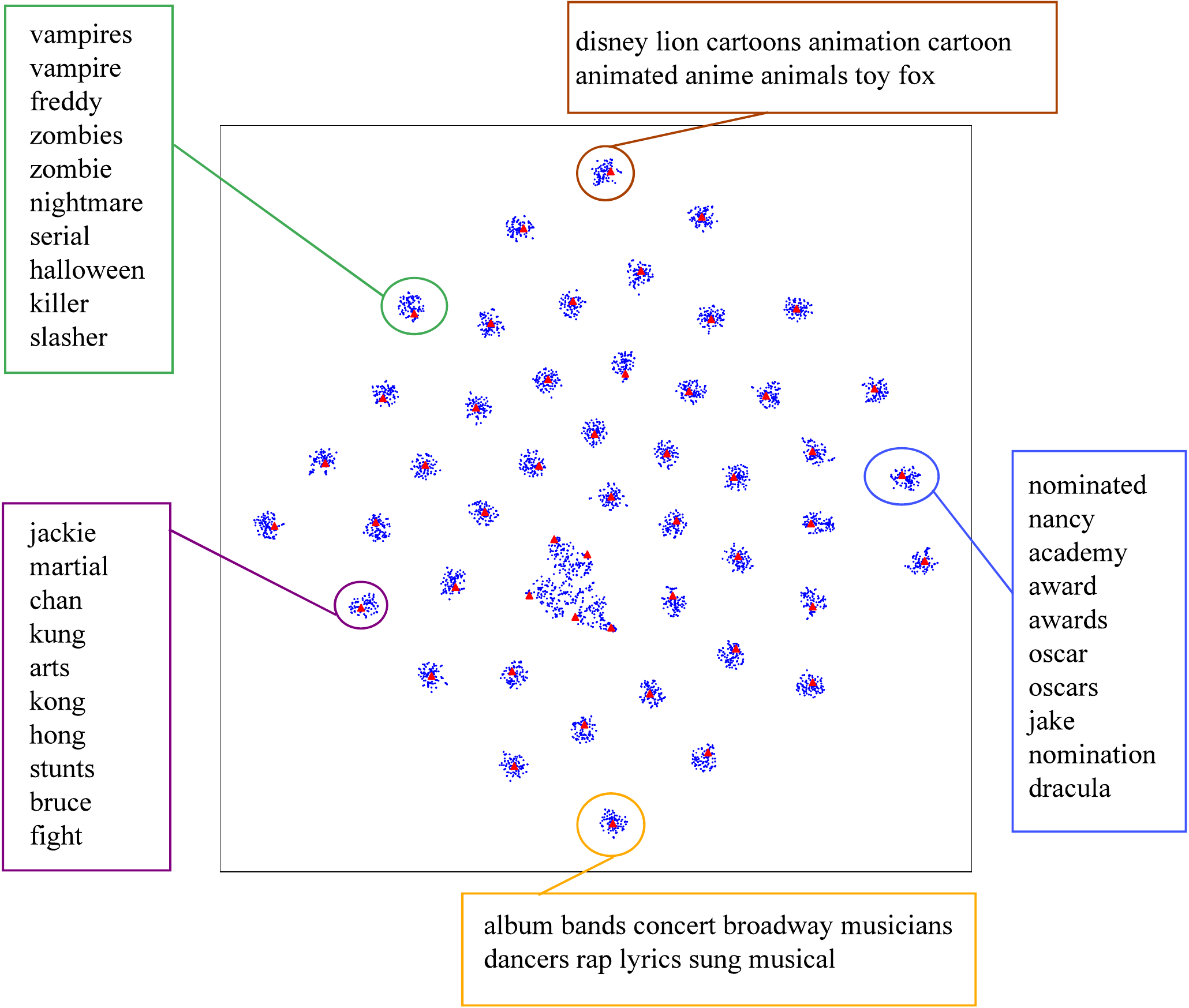}
    \caption{
        Annotations of top words of discovered topics in the semantic space.
    }
    \label{fig_annotation}
\end{figure*}

\begin{figure*}[!ht]
    \centering
    \begin{subfigure}[b]{0.25\linewidth}
        \centering
        \includegraphics[width=0.8\linewidth]{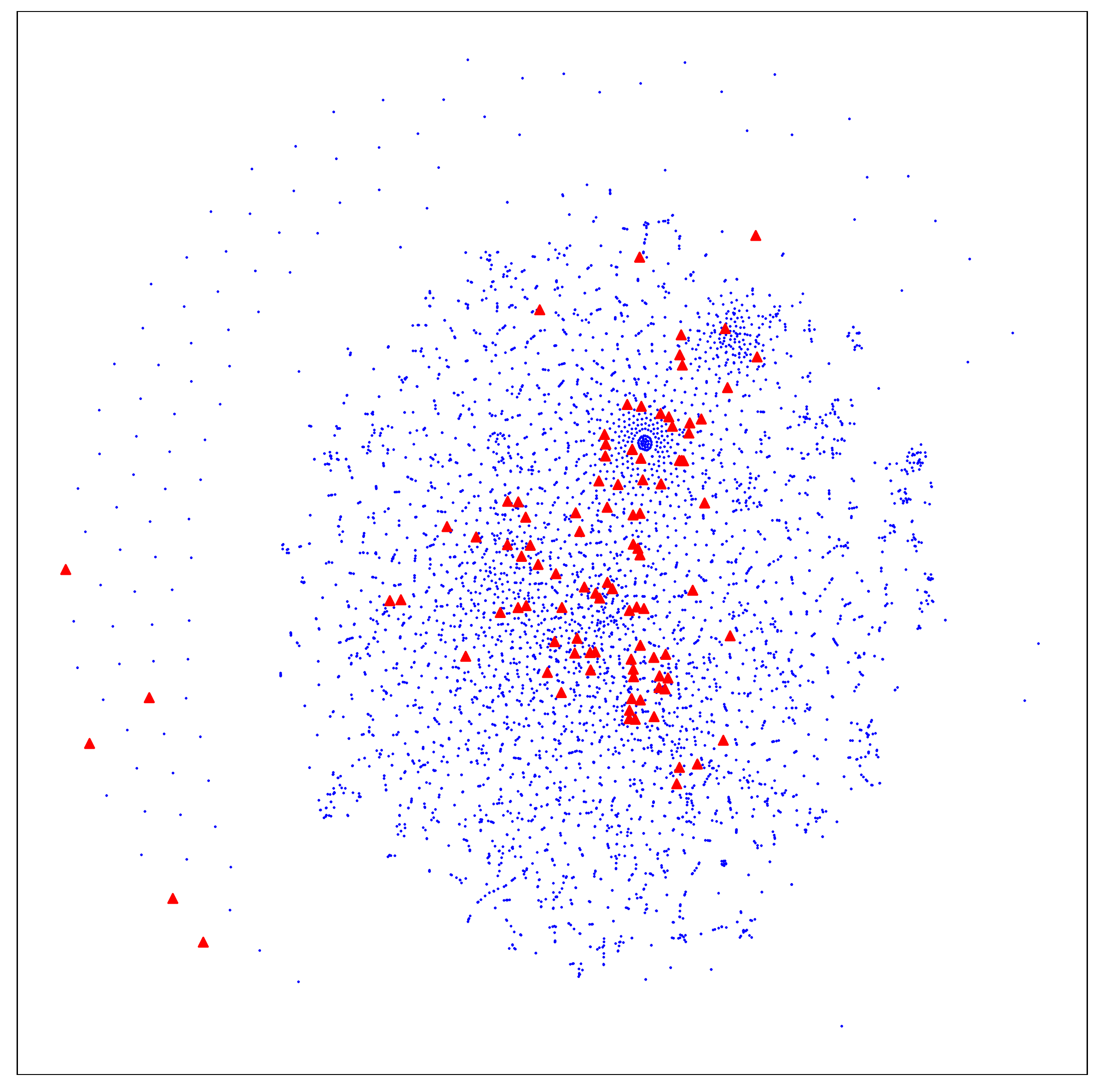}
        \caption{ETM}
    \end{subfigure}%
    \begin{subfigure}[b]{0.25\linewidth}
        \centering
        \includegraphics[width=0.8\linewidth]{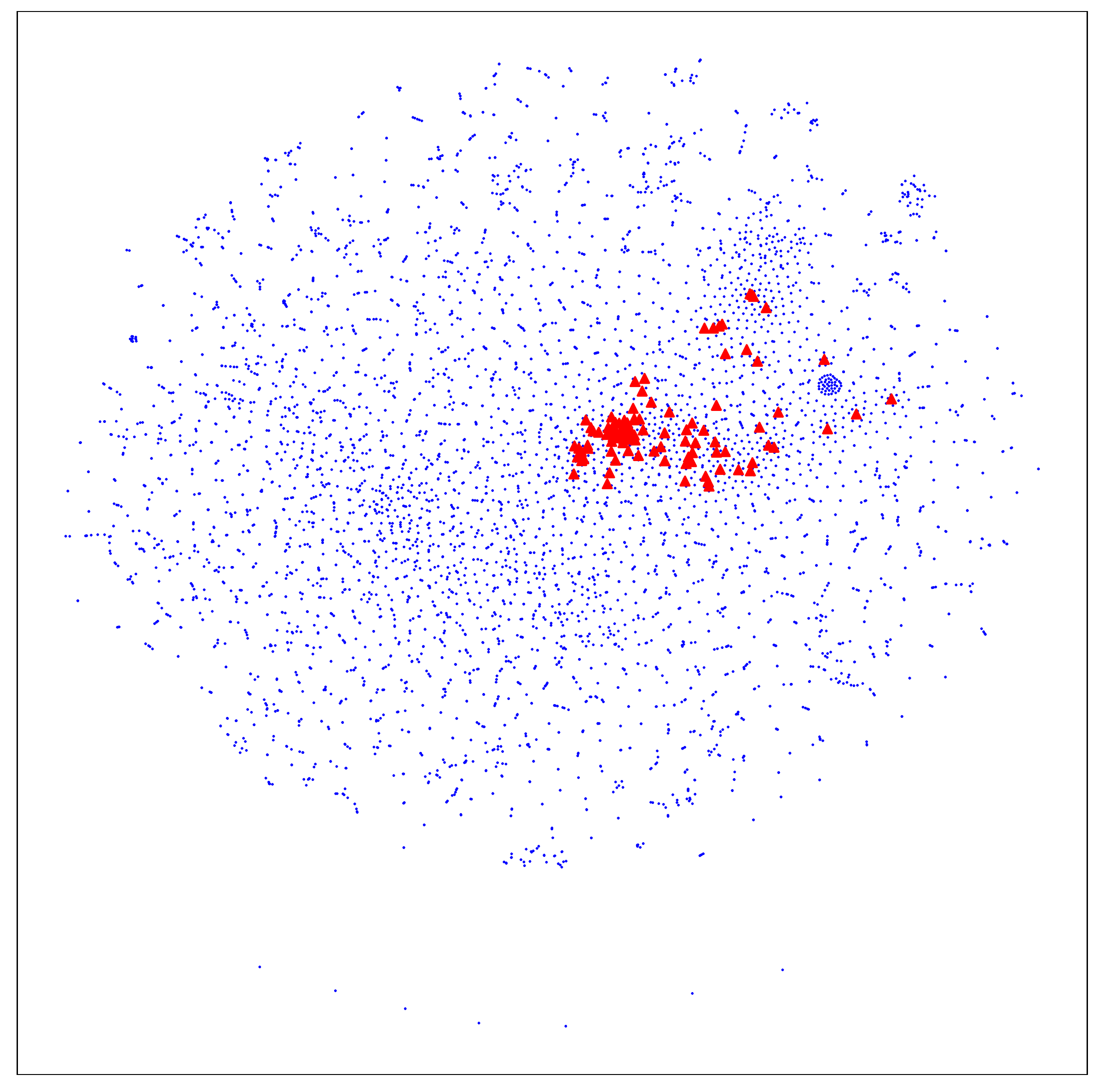}
        \caption{NSTM}
    \end{subfigure}%
    \begin{subfigure}[b]{0.25\linewidth}
        \centering
        \includegraphics[width=0.8\linewidth]{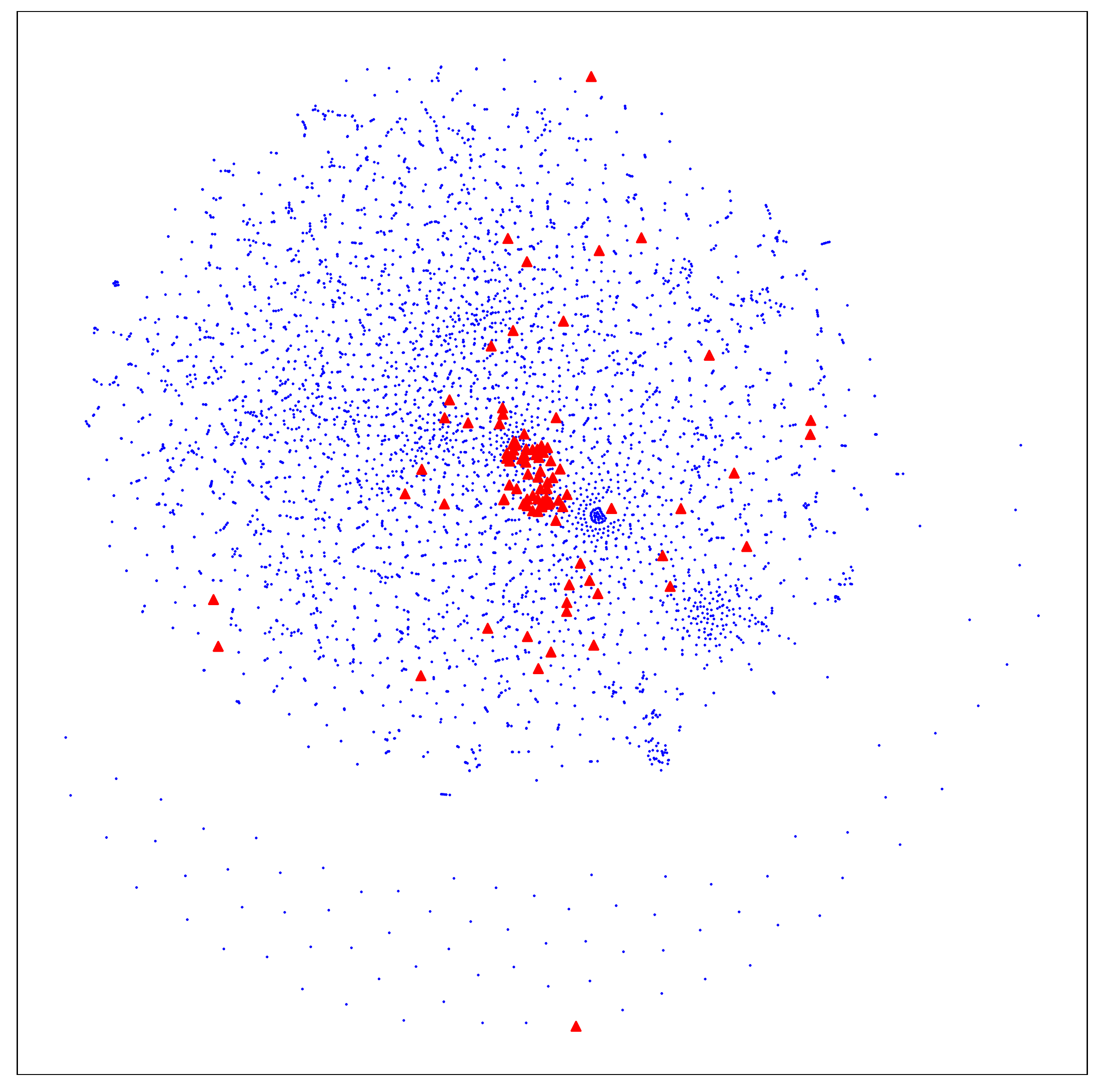}
        \caption{WeTe}
    \end{subfigure}%
    \begin{subfigure}[b]{0.25\linewidth}
        \centering
        \includegraphics[width=0.8\linewidth]{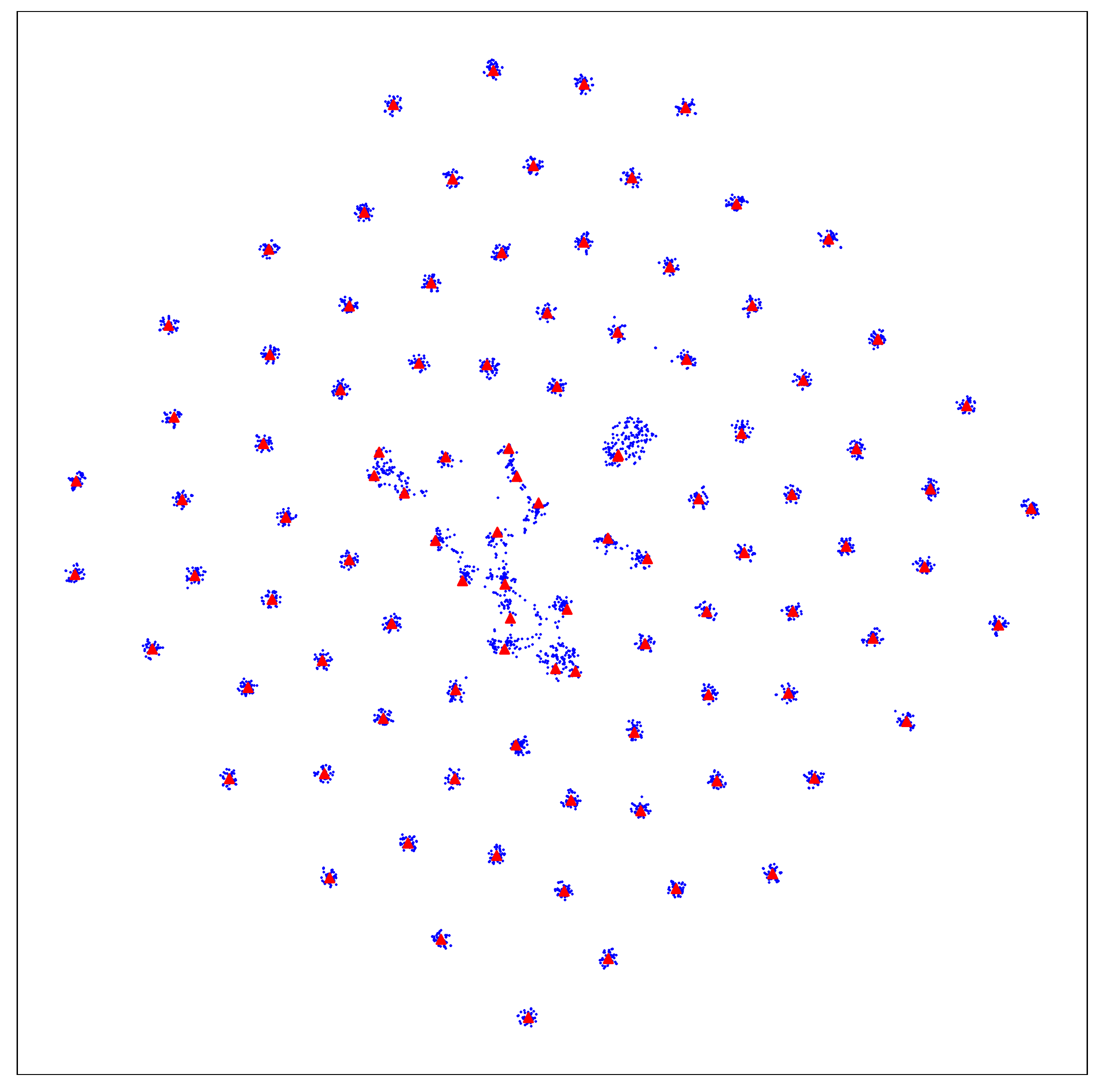}
        \caption{\textbf{\modelname{}}}
    \end{subfigure}%

    \caption{
        t-SNE \citep{Maaten2008} visualization of word embeddings ({\color{blue} $\bullet$}) and topic embeddings ({\color{red} $\blacktriangle$}) \textbf{under 100 topics}.
        Topic embeddings commonly collapse together in state-of-the-art models (ETM \citep{dieng2020topic}, NSTM \citep{zhao2020neural}, and WeTe \citep{wang2022representing}).
        In contrast, \modelname{} can avoid the collapsing by forcing each topic embedding to be the center of a separately aggregated word embedding cluster.
    }
    \label{fig_motivation_K100}
\end{figure*}

\section{Model Implementation} \label{sec_appendix_implement}
    For pre-trained word embeddings, we employ 200-dimensional GloVe \citep{pennington2014glove} \footnote{\url{https://nlp.stanford.edu/projects/glove/}}.
    For the Sinkhorn's algorithm of \modelname{}, we set the maximum number of iterations as 1,000, the stop tolerance 0.005, and $\varepsilon$ 0.05 following \citet{cuturi2013lightspeed}.
    For our \modelname{},
    the prior distribution is specified with Laplace approximation \citep{Hennig2012} to approximate a symmetric Dirichlet prior as $\mu_{0,k} = 0$ and $\Sigma_{0,kk} = (K-1) / (\alpha K)$ with hyperparameter $\alpha$.
    We set $\alpha$ as $1.0$ following \citet{Card2018a}.
    Our encoder network is the same as \citet{Srivastava2017,Wu2020,Wu2020short}: a MLP that has two linear layers with softplus activation function, concatenated with two single layers each for the mean and covariance matrix.
    We use Adam \citep{Kingma2014} to optimize model parameters.
    See other implementation details in our code.

\section{Perplexity Results} \label{sec_appendix_perplexity}
    We report the perplexity results in \Cref{tab_perplexity}.
    Here we do not include some neural topic models (WLDA, NSTM, WeTe) as they are inapplicable to the perplexity approximation with ELBO (See explanations in  \citet{Miao2016,Srivastava2017,Nan2019,zhao2021topic}).
    \Cref{tab_perplexity} shows our \modelname{} also achieves the best perplexity results (lower is better).

\section{Robustness to the Number of Topics} \label{sec_appendix_range}
    Besides the aforementioned results under $K\!\!=\!\!50,100$ (\Cref{tab_topic_quality,tab_clustering}),
    we also experiment under $K\!\!=\!\!10,20,30,40,60,70,80,90$ on 20NG to verify the robustness of our method.
    As shown in \Cref{tab_range_topic_quality,tab_range_clustering},
    we see that \modelname{} consistently outperforms baseline models in terms of both topic quality and clustering.
    These show that the performance improvements of our \modelname{} are robust to the number of topics.

\section{Comparison of Coherence Metrics} \label{sec_appendix_comparison_coherence_metrics}
    \citet{roder2015exploring} have empirically shown that $C_V$ is a better coherence metric which has better consistency with human judgment than traditional metrics like NPMI, UCI, and UMass \cite{bouma2009normalized,chang2009reading,Newman2010,mimno2011optimizing}.
    We also confirm this argument in our experiments: we find NPMI, UCI, and UMass tend to give higher scores to trivial and less informative topics.
    For example,
    we see in \Cref{tab_coherence_metric},
    Topic\#4-6 are more coherent and informative than Topic\#1-3.
    But unfortunately, NPMI, UCI, and UMass give much higher scores to Topic\#1-3 instead of Topic\#4-6.
    In contrast,
    $C_V$ is more reasonable as it gives higher scores to Topic\#4-6.
    Note that it is incorrect to directly compare a NPMI/UCI/UMass score with a $C_V$ score because they are in different scales.

\begin{table*}[!ht]
    \centering
    \setlength{\tabcolsep}{2mm}
    \renewcommand{\arraystretch}{1.1}
    \resizebox{0.95\linewidth}{!}{
        \begin{tabular}{rrrrl}
        \toprule
        NPMI  & UCI   & UMass & $C_V$    & Top words of topics \\
        \midrule
        0.069 & 0.746 & -0.900 & 0.272 & \textbf{Topic\#1:} way come actually make yet example even fact though rather \\
        0.065 & 0.615 & -0.967 & 0.272 & \textbf{Topic\#2:} fact even indeed though kind something way always actually things \\
        0.065 & 0.628 & -3.367 & 0.255 & \textbf{Topic\#3:} really pretty something seem seems quite things nothing thing ridiculous \\
        \midrule
        -0.076 & -2.660 & -8.196 & 0.497 & \textbf{Topic\#4:} vampires vampire freddy zombies zombie nightmare serial halloween killer slasher \\
        -0.075 & -2.678 & -9.032 & 0.476 & \textbf{Topic\#5:} disney lion cartoons animation cartoon animated anime animals toy fox \\
        -0.114 & -3.367 & -6.120 & 0.452 & \textbf{Topic\#6:} jackie martial chan kung arts kong hong stunts bruce fight \\
        \bottomrule
        \end{tabular}%
    }
    \caption{
        Comparison of coherence metrics.
        $C_V$ gives high scores to coherent topics (Topic\#4-6) while traditional metrics (NPMI, UCI, UMass) assigns high scores to less informative topics (Topic\#1-3).
    }
    \label{tab_coherence_metric}%
\end{table*}%

\section{Visualization of Embedding Space} \label{sec_appendix_visual}
    We visualize the learned topic and word embeddings with t-SNE \citep{Maaten2008} \textbf{under 100 topics} (\Cref{fig_motivation} is under 50 topics).
    \Cref{fig_motivation_K100} shows while the topic embeddings mostly collapse together in the state-of-the-art baselines,
    our \modelname{}
    avoids the collapsing of topic embeddings
    by forcing each topic embedding to be the center of a separately aggregated word embedding cluster.
    This illustrates that our \regname{} also works effectively under a larger number of topics.

    We furthermore annotate the semantic embedding space with the top related words of discovered topics by \modelname{} under $K\!\!=\!\!50$ as shown in \Cref{fig_annotation}.
    We see that each word embedding cluster represents a diverse and coherent topic.
    This shows that our \modelname{} effectively clusters the embeddings of coherent words.

\onecolumn

\section{Full Lists of Discovered Topics} \label{sec_appendix_full_list}

Below are the discovered topics of different models from IMDB under 50 topics ($K\!\!=\!\!50$).

\textbf{ETM}	\\
\texttt{
\scriptsize{
Topic\#1:	director etc dialog cult ben joe steve mans todays bruce \\
Topic\#2:	killer police kill killed murder car head shot killing body \\
Topic\#3:	instead rather put mind together despite come beyond without alone \\
Topic\#4:	series first later one time also since made set years \\
Topic\#5:	art rather use special little like sci king bit style \\
Topic\#6:	dark fantasy earth full begin world open end setting set \\
Topic\#7:	family father young wife son mother child life children daughter \\
Topic\#8:	show shows episode watching see watch episodes television star dvd \\
Topic\#9:	boring entertaining pointless silly dull humour pretentious sub unfunny entertained \\
Topic\#10:	like better good especially end look much done way just \\
Topic\#11:	war world history people documentary american life society german soldiers \\
Topic\#12:	film films cinema directors filmed hollywood filmmakers cinematography welles noir \\
Topic\#13:	new one way time world different place two live day \\
Topic\#14:	like just one way made much times really even feel \\
Topic\#15:	music musical song songs dance stage oscar singing voice career \\
Topic\#16:	even one though also far way still made yet better \\
Topic\#17:	part like end also along two short lines now line \\
Topic\#18:	self somewhat sense seems narrative character sexual becomes often person \\
Topic\#19:	story book great perfect gives comes work novel excellent brilliant \\
Topic\#20:	supposed keep get wanted got trying never kept going turn \\
Topic\#21:	people just really know think say maybe something things understand \\
Topic\#22:	john michael james robert richard paul george jack cast played \\
Topic\#23:	scene scenes sex plot slow violence blood boring twist nudity \\
Topic\#24:	action man death gets fight hero back another comes face \\
Topic\#25:	plot original poor pretty unfortunately decent terrible nothing budget completely \\
Topic\#26:	story also makes interesting many stories however quite make style \\
Topic\#27:	great really good see recommend fan liked definitely thought watch \\
Topic\#28:	seen ever one worst years saw see beginning time remember \\
Topic\#29:	one time make give anyone made take get making another \\
Topic\#30:	time minutes just long hour half back night couple looking \\
Topic\#31:	house goes starts gets looks red night tries takes strange \\
Topic\#32:	love life heart beautiful human loved lives wonderful amazing world \\
Topic\#33:	first one ending time two end second just made last \\
Topic\#34:	best role actor performance played also good performances actors play \\
Topic\#35:	characters character plot director dialogue film real drama development acted \\
Topic\#36:	actors cast good acting throughout script moments story film written \\
Topic\#37:	really thing good guy watch movie whole yes want waste \\
Topic\#38:	course much seems audience quite less picture overall bit opinion \\
Topic\#39:	man young men women two woman town small female local \\
Topic\#40:	like high school game video one quality camera sound shot \\
Topic\#41:	bad even acting like just awful made stupid worse crap \\
Topic\#42:	old little girl get kids girls boy kid cool year \\
Topic\#43:	movie movies watch watching actors people theater theaters disaster entertainment \\
Topic\#44:	make see sure say think need really looks know good \\
Topic\#45:	version american truly classic japanese british era french cartoon tale \\
Topic\#46:	funny comedy fun laugh humor jokes watch hilarious comic imagine \\
Topic\#47:	one like around sort looking kind good main look just \\
Topic\#48:	actually real black just white believe mean see either know \\
Topic\#49:	horror effects low dvd english movies films zombie dead genre \\
Topic\#50:	fact reason actually even never simply none however obvious nothing \\
}
}

\textbf{HyperMiner}\\
\texttt{
\scriptsize{
Topic\#1:	jack musical oscar song songs george dance jane mary scott \\
Topic\#2:	just people like think know something see get say want \\
Topic\#3:	effort attempts lacks poor merely intended lacking whilst grace appeal \\
Topic\#4:	one even end way little part character make never plot \\
Topic\#5:	also quite interesting however rather many much although bit though \\
Topic\#6:	one great seen ever best see watch time every enjoy \\
Topic\#7:	even one end way little part character make plot never \\
Topic\#8:	just like people know something think say see want get \\
Topic\#9:	life love beautiful real heart TRUE romantic lives romance dream \\
Topic\#10:	even one end part little character way make never point \\
Topic\#11:	film films made cinema making silent festival director makers filmmakers \\
Topic\#12:	scene get gets guy girl around back away head getting \\
Topic\#13:	movie movies bad made watch make acting watching even plot \\
Topic\#14:	see saw thought watched dvd watching got video felt went \\
Topic\#15:	good like really just better lot much look pretty nice \\
Topic\#16:	bad horror worst awful terrible acting waste budget worse low \\
Topic\#17:	police crime cop gun soldiers gang prison church agent charlie \\
Topic\#18:	even end way character seems make never little part fact \\
Topic\#19:	funny comedy fun kids school laugh humor jokes hilarious christmas \\
Topic\#20:	just people know like think something see say get want \\
Topic\#21:	even seems fact enough plot end least character audience make \\
Topic\#22:	show series shows episode episodes television season pilot trek writers \\
Topic\#23:	car game red dog cat camp van steve chase eye \\
Topic\#24:	one first two new time another second star also world \\
Topic\#25:	effects earth monster sci space battle island match computer adventure \\
Topic\#26:	one great seen ever best see watch time every fan \\
Topic\#27:	man takes find woman help place small comes becomes along \\
Topic\#28:	also many however quite much interesting rather although though bit \\
Topic\#29:	scene get gets around guy girl sex getting back away \\
Topic\#30:	cast role performance actor john excellent play played performances actors \\
Topic\#31:	just like people know think see get something say going \\
Topic\#32:	film films made cinema making silent director independent makers festival \\
Topic\#33:	scenes action dark slow genre fight opening scene sequence violence \\
Topic\#34:	scene get gets guy girl around back sex guys getting \\
Topic\#35:	one first two time new another second also star world \\
Topic\#36:	even end way one little character make plot enough part \\
Topic\#37:	director work camera production music script sound writer shot direction \\
Topic\#38:	human art french world nature reality powerful images experience deep \\
Topic\#39:	war american black white history world country english british documentary \\
Topic\#40:	young family wife father boy son mother children child daughter \\
Topic\#41:	one great seen ever see best watch time every fan \\
Topic\#42:	one great seen ever best see watch time fan every \\
Topic\#43:	old years still now last time year three long back \\
Topic\#44:	death dead evil blood killer house kill night murder killed \\
Topic\#45:	even end little way one plot character enough part make \\
Topic\#46:	lack attempt self premise dull flat fails failed unfortunately pretentious \\
Topic\#47:	story characters original book version read based stories character king \\
Topic\#48:	good like really just better lot much look pretty acting \\
Topic\#49:	cast role performance actor john excellent played play actors performances \\
Topic\#50:	saw see thought dvd watched watching felt got went video \\
}
}

\newpage

\textbf{NSTM} \\
\texttt{
\scriptsize{
Topic\#1:	miller got smith moore just johnson really know davis think \\
Topic\#2:	even one though time way just come fact much make \\
Topic\#3:	know just really going think something come maybe get even \\
Topic\#4:	one another man murder even wanted others death just taken \\
Topic\#5:	movie like monster something actually just movies come thing kind \\
Topic\#6:	one just along now part way another come time around \\
Topic\#7:	just get going come one know even really something way \\
Topic\#8:	one first time also though best another even although came \\
Topic\#9:	sense kind something love really feel thing feeling nothing sort \\
Topic\#10:	movie movies film films just really best like something thing \\
Topic\#11:	just like really come going get even maybe something good \\
Topic\#12:	really movie thing something just things maybe good stuff kind \\
Topic\#13:	interesting movie something funny kind really quite wonderful things fun \\
Topic\#14:	just going really get something maybe come know even thing \\
Topic\#15:	one time first movie best though even just like also \\
Topic\#16:	best one smith moore miller time first just davis james \\
Topic\#17:	movie film just even best though films actually fact really \\
Topic\#18:	just show even come time one good really going know \\
Topic\#19:	wonderful amazing terrific good really fantastic best something thing pretty \\
Topic\#20:	wearing wore wear dress look dressed wears clothes worn shirt \\
Topic\#21:	even just come way going get make time though one \\
Topic\#22:	really something think know thing maybe just things going good \\
Topic\#23:	movie film best fact one though example even story life \\
Topic\#24:	music best musical songs movie like featured one song playing \\
Topic\#25:	really something seems pretty quite things seem thing nothing think \\
Topic\#26:	goes takes tells gets comes finds makes knows everyone happens \\
Topic\#27:	just even really something come going like actually things get \\
Topic\#28:	just one even something come way really like always good \\
Topic\#29:	funny silly movie stuff amusing scary fun hilarious cheesy boring \\
Topic\#30:	even though much make way just come actually fact one \\
Topic\#31:	just even come one way good really though going something \\
Topic\#32:	even fact though way come actually make yet indeed something \\
Topic\#33:	just one even come time way going coming another get \\
Topic\#34:	something really even just things good actually always way kind \\
Topic\#35:	one even time now though come came last another also \\
Topic\#36:	best movie film one good films actor like movies time \\
Topic\#37:	movie film movies films comedy drama best hollywood starring story \\
Topic\#38:	mother daughter wife sister friend husband married couple actress love \\
Topic\#39:	fact even example though way one rather much indeed life \\
Topic\#40:	nose eyes hand just mouth legs fingers neck teeth like \\
Topic\#41:	really just something think going know things maybe even thing \\
Topic\#42:	even come know just fact think something way really actually \\
Topic\#43:	one even just come now time life mother know though \\
Topic\#44:	just like inside get even look come one everything away \\
Topic\#45:	really maybe thing know something think things just going everybody \\
Topic\#46:	one just another came time back now come went got \\
Topic\#47:	fact even something really though always way actually things kind \\
Topic\#48:	time first one just last second next came coming play \\
Topic\#49:	horrible awful terrible horrific thing horrifying frightening kind shocking really \\
Topic\#50:	film movie films movies best directed drama feature picture though \\
}
}

\newpage

\textbf{WeTe} \\
\texttt{
\scriptsize{
Topic\#1:	tonight fifth terrific kings loser wow fourth lucky bang grabs \\
Topic\#2:	australia progress reached secondly environment suspects interests scores continued press \\
Topic\#3:	profanity striking refuse stress sue complain survived fatal contact cracking \\
Topic\#4:	located population neighborhood nation distance owns cox traffic centered bell \\
Topic\#5:	graphics wholly puzzle map chapter attraction medium reader composed edition \\
Topic\#6:	acclaimed exquisite gothic vivid splendid stark sublime lively photographer literary \\
Topic\#7:	theaters cinemas preview trailers ratings extras studios rental par nyc \\
Topic\#8:	journalist calls debate complaint behaviour flawed defense civil charge innocence \\
Topic\#9:	discovering pursuit rid junk disappears cheating pretending petty discovers saving \\
Topic\#10:	john big new star match city james game george stars \\
Topic\#11:	predator heartbreaking terrifying bravo suspenseful intrigue paranoia menace unsettling mafia \\
Topic\#12:	story characters life character people way real sense love much \\
Topic\#13:	pal moody greek trend global corporate depression combat transformed sin \\
Topic\#14:	shaky hung gray shine rough heights staring screens clad casts \\
Topic\#15:	blacks teams merit races thru selection junior earned ranks groups \\
Topic\#16:	interestingly damned awfully astounding unreal incidentally screwed rendered alas instantly \\
Topic\#17:	original version book classic read novel adaptation written king sequel \\
Topic\#18:	adding sticking thread added process easier hang repeating suspend pieces \\
Topic\#19:	recommendation months raped policeman restored retired morning month weeks six \\
Topic\#20:	sorry understands regret lately teenager fond disliked worry unhappy troubles \\
Topic\#21:	show first years series time see saw since still now \\
Topic\#22:	records bollywood pulp futuristic romp bars circus punk hardcore gems \\
Topic\#23:	huh gotta cried swear guessed shouting shake fooled dude yelling \\
Topic\#24:	builds adds reaches agrees teaches introduces marries explains resembles threatens \\
Topic\#25:	rat tank burn duck trees dirt rabbit tree snake burning \\
Topic\#26:	film one films scenes director also time story plot even \\
Topic\#27:	brad ron betty matt dan ryan flynn glover anderson ann \\
Topic\#28:	war american world black white documentary history america people political \\
Topic\#29:	family young kids school father old girl children child mother \\
Topic\#30:	music musical songs song dance voice dancing singing rock stage \\
Topic\#31:	familys anyones hitchcocks wouldve shouldve everyones couldve wifes itll expect \\
Topic\#32:	great best good role cast comedy actor love character actors \\
Topic\#33:	active musician venture learning overcome teaching taught remained accomplished concepts \\
Topic\#34:	sport gross caliber definition mere waves design games rip aforementioned \\
Topic\#35:	women sex girls woman female men violence sexual gay scenes \\
Topic\#36:	killer death murder police thriller cop crime kill mystery michael \\
Topic\#37:	abc holiday aired eve midnight remake introduce mtv broadcast began \\
Topic\#38:	movie movies watch watching acting plot story scenes wifes familys \\
Topic\#39:	just like really bad good get one think see even \\
Topic\#40:	man back gets goes two get takes house around away \\
Topic\#41:	alcoholic wine cruise vegas serving con businessman beverly bent california \\
Topic\#42:	tech korean victory capital temple wwii riot seconds empire dragon \\
Topic\#43:	recorded guitar album tracks reviewer recording blues noted singers sung \\
Topic\#44:	horror action effects special budget evil gore blood low fight \\
Topic\#45:	hoot tarantino bergman dracula buster creator mario olds hack buff \\
Topic\#46:	origin cultural versus conscience roots reunion politics christ divorce dignity \\
Topic\#47:	confident keen tad flair shy lyrics smart impressed nod calm \\
Topic\#48:	shoddy tasteless formulaic imaginable inane unfunny tripe overacting yawn incompetent \\
Topic\#49:	doom beware compassion foul mayhem rides subtlety cue awe wicked \\
Topic\#50:	simultaneously readers mute dimension complexity critical derivative essence perspective account \\
}
}

\newpage

\textbf{DKM} \\
\texttt{
\scriptsize{
Topic\#1:	cooper india family victor town indian dorothy uncle alice jake \\
Topic\#2:	animated animation disney voiced lion bugs batman cartoon cartoons anime \\
Topic\#3:	christmas disney musical songs bill timeless prince art rock loved \\
Topic\#4:	gags gag footage channel jokes television pilot nostalgic smoking comedy \\
Topic\#5:	rock magazine christian chris daddy roger access page jesus school \\
Topic\#6:	jennifer eva comedy jokes genre plot humor sex nicole pie \\
Topic\#7:	christ games jesus religion bible game freeman christian theory vampires \\
Topic\#8:	eddie woody allen murphy comedians comedian keaton funnier sandra williams \\
Topic\#9:	noises father fear heroine summer toys humor adolescent premise trick \\
Topic\#10:	wedding prince grace betty affair sally marry opera secretary marriage \\
Topic\#11:	winner neil star oscar debut hit screenplay jeff ben stars \\
Topic\#12:	politically political racist garbage black wing jokes free foul stereotype \\
Topic\#13:	children gay baby garbage andy virgin parents fox offensive abuse \\
Topic\#14:	jokes jackson lisa predictable recycled wasted murphy writers williams rock \\
Topic\#15:	waste crap worst costs garbage horrible wasting ashamed sucks pile \\
Topic\#16:	erotic sexuality nudity explicit nude porn lesbian sexual sex photos \\
Topic\#17:	seasons season episodes episode abc trek show aired sitcom series \\
Topic\#18:	delightful parker grant gentle comedies henry witty grim delight arthur \\
Topic\#19:	white black dated english clothing costumes costume queen period heroine \\
Topic\#20:	halloween horror freddy scares slasher carpenter eerie haunted gory scary \\
Topic\#21:	cried ned meryl streep charlie cry touched heartbreaking dan warming \\
Topic\#22:	seagal abc jet tripe hardy stan martial heist arts drivel \\
Topic\#23:	andrews noir detective eastwood harry hopper investigation fbi cop criminals \\
Topic\#24:	lynch art imagery noir poetry images waters landscapes poetic symbolic \\
Topic\#25:	moore female soap arthur daughter island butler copy nurse pregnant \\
Topic\#26:	laugh sam winner comedy loser cop funniest golden hilarious simon \\
Topic\#27:	bollywood indian india hudson khan soap dialogs indians dialogues opera \\
Topic\#28:	santa homeless andrews truck airport car husband security mom wakes \\
Topic\#29:	spike match angle baseball tag indians hudson stewart ring flynn \\
Topic\#30:	sinatra musicals kelly powell mgm broadway musical dance numbers gene \\
Topic\#31:	christmas santa childrens holiday betty age ann adult children toy \\
Topic\#32:	robin don batman hood villains burt pacino dick delivery adam \\
Topic\#33:	adaptation emma jane adaptations timothy bbc novel kenneth book bates \\
Topic\#34:	sons philip son daniel father mother jewish hoffman norman davis \\
Topic\#35:	cars rent scared channels vhs lol expecting cage renting cried \\
Topic\#36:	woody disjointed pacing distracting uninteresting development leonard tension subplots hour \\
Topic\#37:	trek alien planet superman science scientists invisible aliens outer scientist \\
Topic\#38:	fantasy christmas magic effects magical santa special holiday childrens child \\
Topic\#39:	gay blah gray spike indulgent lesbian jerk insulting rape racist \\
Topic\#40:	martial arts kung jet hong chan kong ninja jackie sword \\
Topic\#41:	zombies zombie werewolf vampires monster creature dinosaur rubber shark topless \\
Topic\#42:	indie chaplin festival budget pleasantly independent jake distribution spike victor \\
Topic\#43:	computer plot imaginative tech sub slasher budget holes video jennifer \\
Topic\#44:	low code monkey rental store computer scientist laughs budget executive \\
Topic\#45:	drew daughter adult father adults sports parents younger teen kids \\
Topic\#46:	propaganda jews hitler germans nazis soviet nazi civil americans jewish \\
Topic\#47:	baby jokes daughter humor amusing unfunny silly cute comedy funny \\
Topic\#48:	bands metal album punk kids tap band santa nerd school \\
Topic\#49:	bollywood manager eddie rap van store dance tap singer choreography \\
Topic\#50:	eddie album nominated bands concert vhs murphy jackie awards oscars \\
}
}

\newpage

\textbf{DKM+Entropy}\\
\texttt{
\scriptsize{
Topic\#1:	zone european even choppy showing shows evening shy china side \\
Topic\#2:	sincerely workers warn thank yes ever seen old way semi \\
Topic\#3:	finally morgan drug tape tap tank drugs remember remind reminded \\
Topic\#4:	laughter distracting discussion social disbelief disappoint moronic motivation dinosaurs dimension \\
Topic\#5:	bin camera hollywoods couldve amazingly ambition pacing paced titled awe \\
Topic\#6:	reviewers spend spoke got split gore gordon motion gone spike \\
Topic\#7:	expected opened entry sea canada june north installment ticket april \\
Topic\#8:	zone period overcome overlooked pace pacino discovering painful painfully paper \\
Topic\#9:	naked cousin raped fat neo crazed fans twins indian indians \\
Topic\#10:	abandoned tons discovers spoiled died die diane rat near nearby \\
Topic\#11:	clothing fitting trained amazingly robot borrowed masterpiece primarily heavily comic \\
Topic\#12:	overwhelming disjointed heartfelt flair understandable captures sincere angst spirited dreck \\
Topic\#13:	buff dude like saw survivor genuine start estate saving ever \\
Topic\#14:	physical day semi ever way old due funny seen harm \\
Topic\#15:	zone insipid interestingly interesting interest single intentions intensity intelligence insulting \\
Topic\#16:	roof practically underneath tame colour albeit colors shorter color badly \\
Topic\#17:	episode identify corruption costs project ignored promise ignore couple couples \\
Topic\#18:	getting spring dollars stevens steven message stephen step doctors mickey \\
Topic\#19:	rip dogs images discovered chinese illegal objects baseball possibly display \\
Topic\#20:	picked patients sandra hour hours howard boat human continually sarah \\
Topic\#21:	zone brilliance principal prior brief describing prize description process produced \\
Topic\#22:	urge decision rushed agree navy agreed agrees ahead aid defense \\
Topic\#23:	won japan spain union april italy south bank north entry \\
Topic\#24:	homeless womens disturbed helps cares compassion souls incompetent caring ashamed \\
Topic\#25:	episode identify corruption costs project ignored promise ignore couple couples \\
Topic\#26:	improved lacked relative reflection reflect stress corporate respect lighting supporting \\
Topic\#27:	zone insipid interestingly interesting interest single intentions intensity intelligence insulting \\
Topic\#28:	checked growing cole come comes hal gun guessed study comments \\
Topic\#29:	rocket rescue slice mates burns strong stronger strongest bus respective \\
Topic\#30:	semi seen due old way physical day ever funny sentimental \\
Topic\#31:	benefit buying paid guarantee suits patricia wealthy pay paying pays \\
Topic\#32:	cult japanese killers sympathy kill kevin justin josh joins jimmy \\
Topic\#33:	zone playing physical picking piece battles credited place places plane \\
Topic\#34:	canada lower april june entry north installment sea ticket opened \\
Topic\#35:	struggle americans note notch barry barrel nostalgic task screen tape \\
Topic\#36:	funny day physical semi ever way old due seen zone \\
Topic\#37:	zone place phil philosophical philosophy class claire phrase physical claims \\
Topic\#38:	invasion army wwii opened june bank join third next fourth \\
Topic\#39:	way day ever old due seen funny physical semi person \\
Topic\#40:	positive crisis elements painful serial television effects seen two causes \\
Topic\#41:	gas plenty receives dislike average regarding treatment household safety concern \\
Topic\#42:	zone honesty holy hollywood holiday holds shared hoffman hitting hits \\
Topic\#43:	system college supported identical skip trial mentioned furthermore actual half \\
Topic\#44:	funny ever day old seen way physical due semi relationship \\
Topic\#45:	funny ever day semi seen physical way old due psychological \\
Topic\#46:	stuff cried intentionally technical innocence anyway tender anything anyone incident \\
Topic\#47:	compared conviction howard ted technology sci human officers office hundred \\
Topic\#48:	tight understand positive came television seen standard win like semi \\
Topic\#49:	ever semi due old funny physical day way seen solution \\
Topic\#50:	nominated pass nod nomination suspect willing joined warner jean stanley \\
}
}

\newpage

\textbf{ECRTM}\\
\texttt{
\scriptsize{
Topic\#1:	students school teacher student sean high specially class texas shelf \\
Topic\#2:	disney lion cartoons animation cartoon animated anime animals toy fox \\
Topic\#3:	dimensional merely clumsy unpleasant draw potentially consistent handed develop wholly \\
Topic\#4:	santa christmas children kids adults adult child parents relax age \\
Topic\#5:	stories york season match lifetime davis tony currently respected episodes \\
Topic\#6:	wars alien burton dinosaurs outer graphics trek futuristic aliens sci \\
Topic\#7:	terrible costs horrible renting sucks awful avoid rented sounded rent \\
Topic\#8:	budget low values stinks producing violence gratuitous blast spare lighting \\
Topic\#9:	seasons abc episode aired season show episodes network program television \\
Topic\#10:	eastwood andrews hopper fbi westerns investigation policeman crime investigating showdown \\
Topic\#11:	worst superman hats ever les shorter guitar banned maniac mates \\
Topic\#12:	funniest comedies comedy laugh laughing dan black mario white jokes \\
Topic\#13:	japanese japan russian wrestling reynolds kim biased marketing industry vincent \\
Topic\#14:	album bands concert broadway musicians dancers rap lyrics sung musical \\
Topic\#15:	bates adams purchased ann library lasted listed quinn map native \\
Topic\#16:	jackie martial chan kung arts kong hong stunts bruce fight \\
Topic\#17:	cars swedish car thumbs files cop airport sappy jet tomorrow \\
Topic\#18:	bat stuart angela shouldve flynn hoot plague trailer abysmal bath \\
Topic\#19:	sucked crappy stupid twins monkey stupidity darn horrid idiotic puppet \\
Topic\#20:	vhs bought copy tape rental dvd store dvds video cable \\
Topic\#21:	dad son jackson jack father hotel god saved king thinks \\
Topic\#22:	games game bond victor chris germany reunion mexico french marie \\
Topic\#23:	martha familys dysfunctional cowboy cope illness daniel bergman financial property \\
Topic\#24:	titanic waste spike flop dreck advise someday blockbuster junk taylor \\
Topic\#25:	grim page gentle timothy understated magnificent captures poignant passionate debut \\
Topic\#26:	development unlikeable drew roy main pitt character implausible descent believable \\
Topic\#27:	sequel remake original beginning credits improved van scene missed spoilers \\
Topic\#28:	ted lou chaplin bridges mel warren russell sandra butler elizabeth \\
Topic\#29:	football teams randy kicked jeff ensues airplane thugs roof bus \\
Topic\#30:	woody walken jerry kelly hanks allen sinatra tom dances musicals \\
Topic\#31:	lynch ireland bang unusual worthwhile rabbit tops twist quotes temple \\
Topic\#32:	excited waters comments reading expectations yelling book urge disturbed sticking \\
Topic\#33:	excellent fantastic recommend performance highly job brilliant amazing enjoyed definitely \\
Topic\#34:	vampires vampire freddy zombies zombie nightmare serial halloween killer slasher \\
Topic\#35:	christ christian religious faith theory interviews media holy intellectual studying \\
Topic\#36:	jane emma novel adaptation novels book version versions faithful books \\
Topic\#37:	freeman justin nicole holly carrie annie austin glover btw matthew \\
Topic\#38:	eddie murphy unfunny funnier tacky clown humour yawn gags tripe \\
Topic\#39:	artistic medium landscape breathtaking painting imaginative poetry technique contemporary movements \\
Topic\#40:	demons carpenter spooky eerie sleaze myers gothic karen hammer patients \\
Topic\#41:	rubber cabin ninja rat barrel cave corpses splatter lake predator \\
Topic\#42:	festival indie harry welles films buffs film stan lucy gay \\
Topic\#43:	mean maybe sorry hate honestly understand saying else noises someone \\
Topic\#44:	hitler germans soviet civil war wwii russia fought union holocaust \\
Topic\#45:	half improvement stretch ford hour limit respectable swimming covers portion \\
Topic\#46:	heaven cage bravo segment earth robot vegas science planet drove \\
Topic\#47:	wasting existent amateur mtv choppy lowest asleep shaky lifeless stores \\
Topic\#48:	nominated nancy academy award awards oscar oscars jake nomination dracula \\
Topic\#49:	virgin women andy sex sarah male diane kate woman boyfriend \\
Topic\#50:	touched bollywood sadness cried joy feelings emotion inspiring relationships warm \\
}
}

\end{document}